\documentclass[11pt]{article}

\RequirePackage{amsthm,amsmath,amsfonts,amssymb}
\RequirePackage[authoryear,round]{natbib}%% uncomment this for author-year citations

\usepackage{fullpage}
\usepackage[skip=2mm,indent=5mm]{parskip}

\usepackage[pdftex]{hyperref}
\hypersetup{
    unicode=false,          % non-Latin characters in AcrobatÂ¿s bookmarks
    pdftoolbar=true,        % show Acrobat toolbar?
    pdfmenubar=true,        % show Acrobat menu?
    pdffitwindow=false,      % page fit to window when opened
    pdfnewwindow=true,      % links in new window
    colorlinks=true,       % false: boxed links; true: colored links
    linkcolor=DarkBlue,          % color of internal links
    citecolor=DarkGreen,        % color of links to bibliography
    filecolor=DarkRed,      % color of file links
    urlcolor=DarkBlue,          % color of external links
    pdftitle={Correct Looks Better: Pairwise Comparisons Reveal Accuracy Rankings},
    pdfauthor={Mina Remeli and Moritz Hardt},
}

\usepackage{graphicx}
\usepackage{caption}
\usepackage{subcaption}
\usepackage{algorithm}
\usepackage{algpseudocode}
\usepackage[table,xcdraw]{xcolor}
\usepackage{float}
\usepackage{booktabs} 
\usepackage{hyperref}
\usepackage{listings}
\usepackage{tablefootnote}
\usepackage{booktabs}
\usepackage{subcaption}

\usepackage{xcolor}
\definecolor{customred}{HTML}{d62728}

\usepackage{color}
\definecolor{DarkGreen}{rgb}{0.075,0.375,0.075}
\definecolor{DarkRed}{rgb}{0.5,0.1,0.1}
\definecolor{DarkBlue}{rgb}{0.1,0.1,0.5}
\definecolor{Gray}{rgb}{0.2,0.2,0.2}
\newcommand{\block}[1]{%
    \raisebox{\dimexpr(\fontcharht\font`X-1em)/2}{\rule{1em}{#1\dimexpr1em/8}}%
}
\DeclareUnicodeCharacter{2581}{\block{1}}
\DeclareUnicodeCharacter{03C5}{$\upsilon$}

\newcommand{\mnote}[1]{}
\newcommand{\minote}[1]{}

\newcommand{\figwidth}{\linewidth}
\newlength{\figwidthnarrow}
\AtBeginDocument{\setlength{\figwidthnarrow}{\linewidth}}
\renewcommand{\figwidth}{0.65\linewidth}
\AtBeginDocument{\setlength{\figwidthnarrow}{0.6\linewidth}}

\title{Correct Looks Better: Pairwise Comparisons Reveal Accuracy Rankings}

\usepackage{authblk}
\stepcounter{footnote}
\addtocounter{footnote}{-1}
\author[1]{Mina Remeli\thanks{Corresponding author: mina.remeli@tuebingen.mpg.de}}
\author[1]{Moritz Hardt}
\affil[1]{Max Planck Institute for Intelligent Systems, Tübingen, Germany, Tübingen AI Center}
\date{}                     %% if you don't need date to appear
\begin{document}

\maketitle

\begin{abstract}
Pairwise comparisons combined with aggregation methods like Elo have become central to evaluating generative models, yet concerns remain that they reward superficial stylistic cues or display judge biases.
In a more positive turn, we show that model rankings from pairwise comparisons strongly agree with ground-truth-based accuracy rankings when such ground truth is available for comparison. 
By converting five well-known benchmarks into free-form generative evaluations, we find that Elo rankings achieve a Spearman correlation above 0.9 with accuracy rankings and substantially outperform direct evaluation when the judge is weak. 
Furthermore, style and judge bias have only minor effects on model rankings, despite most judgments occurring on pairs where both candidate answers are correct (or incorrect).
On such pairs, we find that repetition after the final answer (\textit{echo}) is a causal driver of judge preference.
\end{abstract} 

\section{Introduction}
\begin{figure}
    \centering
    \includegraphics[width=\figwidthnarrow]{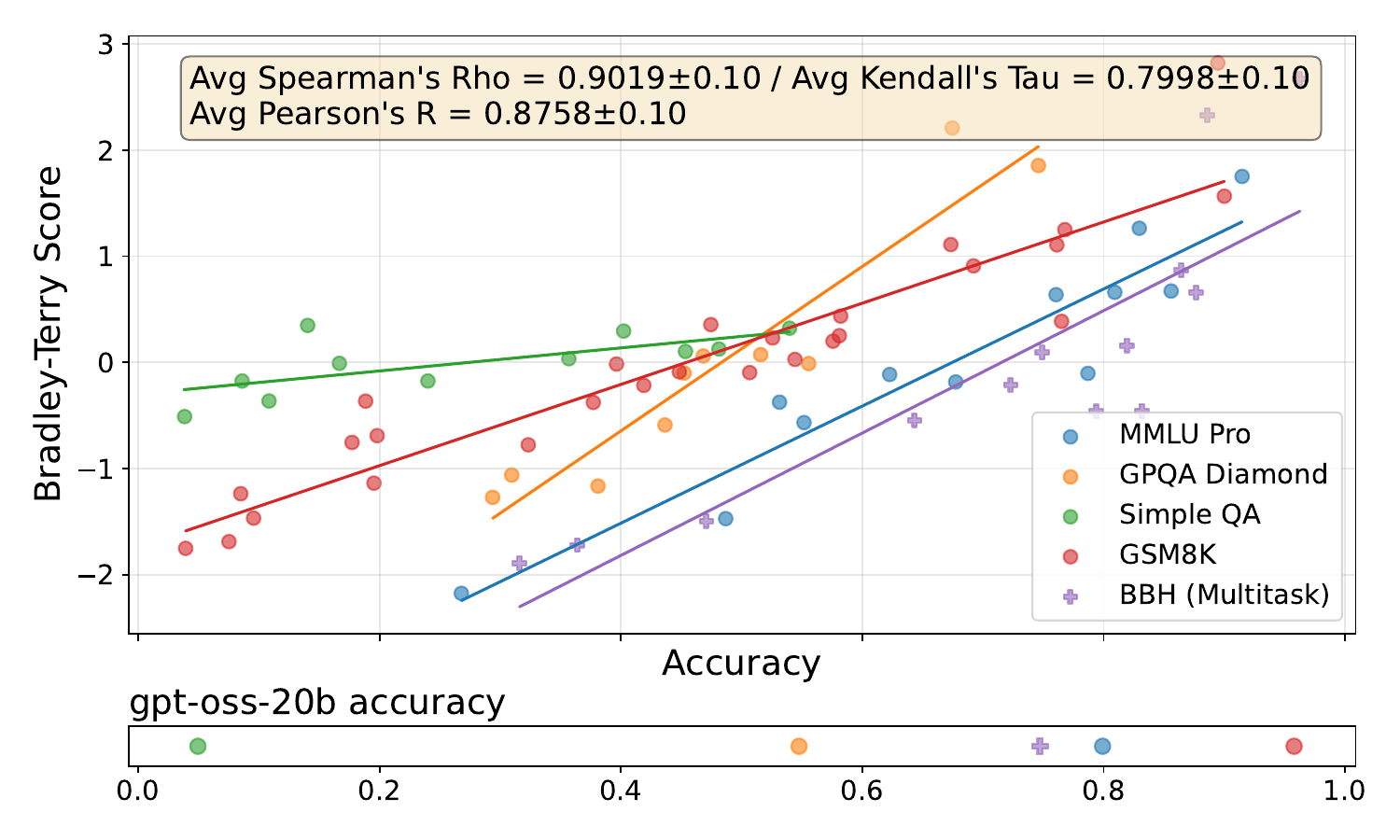}
    \caption{\textbf{High correlation between accuracy and Bradley-Terry scores.} Each point represents a model evaluated on a specific benchmark: we plot accuracy on the x axis and the Bradley-Terry score on the y axis. Bradley-Terry estimates each model's `strength' by aggregating over preferences between two candidate answers (\textit{pairwise comparisons}). We collected pairwise comparisons using \texttt{gpt-oss-20b}. We observe both \textit{high rank correlation} and \textit{high linear correlation} across all benchmarks.}
    \label{fig:main}
\end{figure}

\begin{figure*}[ht]
    \centering
    \includegraphics[width=\linewidth]{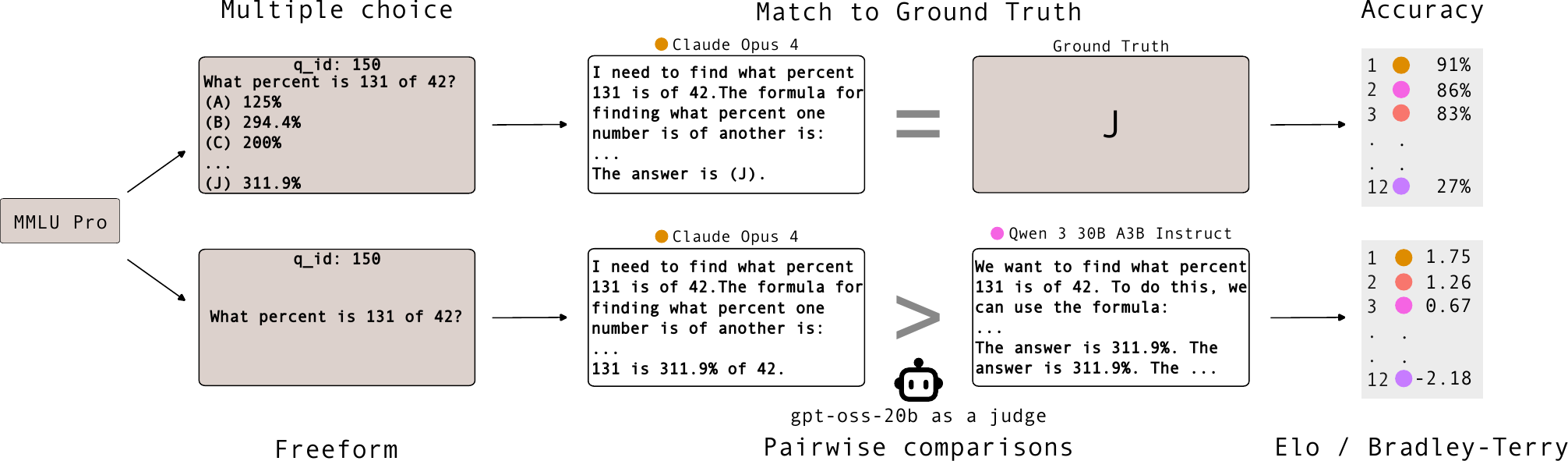}
    \caption{\textbf{From benchmarks to rankings.} For each benchmark we obtain an Elo-style ranking and compare it to the accuracy based ranking. Accuracies are obtained by grading answers based on the ground truth answer (top part of the figure), while Elo-like scores are obtained by aggregating over pairwise comparisons across answer pairs (bottom part of the figure). For multiple choice benchmarks (such as MMLU Pro) we remove the answer options to make them freeform.}
    \label{fig:pipeline}
\end{figure*}

Pairwise comparisons by humans or judge models and aggregated via Elo or Bradley-Terry have become a central tool to evaluate generative models \cite{zheng2023judging, chiang2024chatbot}. They are attractive because they are often simple to collect, broadly applicable across diverse tasks — most prominently in open-ended settings such as Chatbot Arena \cite{chiang2024chatbot} — and naturally yield relative rankings even when absolute measurement is hard. At the same time, there’s been much debate about the validity of pairwise comparisons: they might track surface-level style \cite{Li_Angelopoulos_Chiang_2024}, exploit presentation artifacts \cite{dubois2024length, chen_humans_2024}, or reflect idiosyncratic judge biases \cite{zheng2023judging, wataoka_self-preference_2025}. In short, preferred answers might simply \emph{look better} without revealing correctness or quality. These concerns are especially acute in settings where ground truth is unavailable or expensive, and when the judge may be substantially weaker than the models under comparison.
Validating these concerns, however, requires a controlled comparison — which is only possible when ground truth is available. In this work we ask a basic but largely unanswered question: when ground truth \emph{is} available, do pairwise-preference rankings actually reflect accuracy? To test this directly, we convert established benchmarks with ground truth into free-form generative evaluation protocols and score models using Elo-style rankings derived from pairwise comparisons. Across five well-known benchmarks we find that preference-based rankings align surprisingly well with accuracy-based rankings. In particular, Bradley-Terry rankings achieve high agreement with accuracy, with Spearman rank correlation consistently above 0.9. This suggests that, at least for these tasks, pairwise comparisons recover essentially the same ordering one would obtain from labeled evaluation.
The results are most striking in the setting where pairwise comparisons are most needed: the \emph{evaluation frontier}, where the judge model is far from perfectly solving the task and possibly weaker than the candidate models. In this regime, directly applying the weak judge to determine correctness fails, leading to rankings far from the ground truth accuracy ranking. In contrast, pairwise-preference rankings from weak judges remain well-aligned with the accuracy ranking. These findings have an almost paradoxical implication: we can rank models by accuracy without knowing the models’ accuracies. Resolving this counterintuitive finding, we argue that style and bias can indeed impede \emph{absolute} measurement from pairwise comparisons, but their effect on \emph{rankings} is far more limited.

In more detail, we make the following main observations:
\begin{description}
    \item[Preference rankings agree with accuracy.] We observe both high rank correlation (Spearman's $\rho > 0.9$) and high linear correlation ($R^2 > 0.87$) between the preference-based ranking and the ground-truth accuracy ranking. We tested this on five well-established benchmarks: four single-task (MMLU Pro, GPQA Diamond, SimpleQA, GSM8K) and one multi-task benchmark (BBH). We ranked between 10--27 models depending on the benchmark. 
    \item[Preference rankings work with weak judges.] When the judge model performs poorly on the task, pairwise comparisons lead to substantially better alignment with accuracy than directly asking the judge about correctness. This is especially relevant at the evaluation frontier, where we consider difficult tasks that the judge cannot reliably solve.
    
    \item[Rankings are robust to style and bias.] 
    Style and self-preference bias are the two most commonly cited sources of judge bias in the literature. This is especially concerning given that almost 60\% of pairwise comparisons are \textit{non-discriminative} — pairs where both answers are correct or both are incorrect, leaving superficial cues as the primary available signal. Yet we show that correcting for these biases has only a minor effect on rankings.

    \item[Echo drives judgment on non-discriminative pairs] 
    Surprisingly, rankings continue to align well with accuracy even when restricting to non-discriminative pairs. The model with the higher accuracy tends to win the comparison, despite both answers being correct (or incorrect). We identify \textit{echo} — repetition after the final answer — as a causal driver of judge preference on such pairs.
\end{description}

\section{Related Work}
Pairwise comparisons have long played a central role in the evaluation of NLP systems. The practice dates back to machine translation evaluation \cite{sakaguchi2014efficient}, where they used human preferences over translations to rank models. More recently, Chatbot Arena \cite{chiang2024chatbot} has emerged as a crowd-sourced ranking of LLMs. 
However evaluations from language model judges offer a scalable alternative and are becoming increasingly popular due to their high agreement with human judgment \cite{zheng2023judging}.
A flurry of related work has emerged in an attempt to reproduce Chatbot Arena's human preference based ranking. Works such as LC-AlpacaEval \cite{dubois2024length} and Arena-Hard \cite{li2024crowdsourced} collect pairwise comparisons using LLMs, which are then often aggregated using a rank aggregation algorithm such as Elo \cite{elo1978rating} or Bradley-Terry \cite{bradley1952rank}. 
However, LLM-based pairwise comparisons have been shown to exhibit various types of biases such as length-bias \cite{dubois2024length}, presentation bias \cite{chen_humans_2024} and preferring answers from itself \cite{zheng2023judging, wataoka_self-preference_2025}. 
While related work has focused on \textit{human preference based ranking} of models, our work aims to reproduce the \textit{accuracy} based ranking using pairwise comparisons.

Another related topic is the automated evaluation of LLMs. 
Recently \cite{kelley2025tournament} proposed a tournament-style evaluation framework that combines pre-existing benchmarks with rank aggregation algorithms. They use EleutherAI's Evaluation Harness to determine the winner of each pairwise comparison. Their approach relies on the availability of ground truth labels which makes the reported correlation with accuracy-based ranking less surprising. Our work on the other hand studies automatic evaluation \textit{without access to ground truth}, and uses LLMs as judges to collect pairwise comparisons.
\citet{zheng2023judging} have pioneered pairwise comparisons and LLM-as-a-judge approaches in the freeform generative evaluation setting. Other related work uses judges to automatically match semantically equivalent answers \cite{chandak2025answer}, using CoT reasoning to evaluate generated text \cite{liu_g-eval_2023} or fine-tuning LLMs to evaluate instruction following tasks \cite{wang_pandalm_2024}. These methods  often provide LLM judges with grading rubrics \cite{hashemi_llm-rubric_2024} or incentivize truthfulness through the use of proper scoring rules \cite{wu_elicitationgpt_2025}.

\section{Method}
We study Elo rankings in a freeform generative evaluation setting, with no access to the ground truth. Our goal is to see whether rankings obtained using pairwise comparisons align with the accuracy based ranking on benchmarks in a standard evaluation setting.  We illustrate our general pipeline in Figure \ref{fig:pipeline} with MMLU Pro as an example benchmark.
The source code is publicly available at: \url{https://github.com/socialfoundations/correct-looks-better}

\subsection{Standard evaluation setting}
By standard evaluation setting, we mean a general benchmarking pipeline where the language models are evaluated based on some ground truth answer. To make evaluation easier, benchmarks often enforce some structure that models are required to follow, such as multiple choice questions, fill-in-the-blank, etc. We illustrate this setting in the top part of Figure \ref{fig:pipeline}.

\subsubsection{Accuracy based ranking}
To calculate the accuracy, we followed each benchmark's grading instructions. For most benchmarks, this meant we  extracted the relevant parts of the model answer (usually the string after ``The answer is'') and checked whether it matched the ground truth. For two benchmarks, we used LLM assisted grading to classify answers. We used SimpleQA's \cite{wei2024measuring} template for the grading instructions. For our multitask benchmark, we calculate the micro-averaged accuracy. See Table \ref{tab:bench-summary} for a summary on the accuracy based ranking for each benchmark. The accuracy based ranking serves as our `golden standard'.

\begin{table}[t]
\centering
\caption{\textbf{Accuracy based ranking on benchmarks.} We specify the type of output the benchmark expects, evaluation setting and method of calculating the accuracy. We used \texttt{gpt-oss-120b} for LLM assisted grading. GT stands for ground truth.}
\label{tab:bench-summary}
\resizebox{\figwidth}{!}{%
\begin{tabular}{@{}lccc@{}}
\toprule
\textbf{Benchmark} & \textbf{Type}   & \textbf{Setting} & \textbf{Accuracy} \\ \midrule
MMLU Pro           & multiple choice & zero-shot, CoT   & Match to GT       \\
GPQA Diamond       & multiple choice & zero-shot        & Match to GT       \\
Simple QA          & free-form       & zero-shot        & LLM Grader (w/GT)\tablefootnote{For SimpleQA, we report the ``Correct given attempted'' accuracy, like in \citet{wei2024measuring}.} \\
GSM8K              & free-form       & zero-shot        & LLM Grader (w/GT) \\
BBH (Multitask)    & both            & few-shot, CoT    & Match to GT       \\ \bottomrule
\end{tabular}%
}
\end{table}

\subsection{Freeform generative evaluation setting}
There are two key differences to our previous setting: questions are asked in a freeform format, and we have no access to the ground truth (see bottom part of Figure \ref{fig:pipeline}). While most of the tasks we evaluated on were freeform, some benchmarks use a multiple choice question format. We convert them into freeform by removing the answer options. This allows us to study potential judge biases, which would not be possible for single-letter answers (without CoT / reasoning traces) because there are little to no stylistic cues in the multiple choice setting.

% The output of multiple choice questions is often a single letter (without CoT / reasoning traces), which makes it hard to evaluate whether the model chooses an answer based on stylistic cues rather than the correctness of the answer (because there are little to no stylistic cues in the multiple choice setting). We deliberately convert multiple choice questions into freeform format to study judge biases.

\subsubsection{Elo ranking}
Here we describe how we obtain the Elo style ranking on a given benchmark. Our approach can be broken down into three steps: (i) answer collection to freeform questions, (ii) collecting pairwise comparisons and finally (iii) using an aggregation algorithm (such as Elo or Bradley Terry) to obtain the final model scores. We use Bradley-Terry rather than Elo scoring because it provides the maximum likelihood estimate of model strengths given the full dataset, whereas Elo is an online approximation that processes comparisons sequentially. We compared Bradley-Terry against WinRate, Elo, and TrueSkill and found that it achieves the highest mean rank correlation across benchmarks (Appendix~\ref{app:rank_agg}). 

We begin by generating answers to freeform questions from the models we are trying to evaluate. Next, we collect pairwise comparisons. For each question $q \in Q$, we randomly sample pairs of model answers $a_i, a_j \in A_{q}$. We prompt a large language model to express its binary preference over such pairs:

\begin{quote}
\small\ttfamily
You will be shown two answers to the same question. The answers to the question may or may not be correct. You have to make a judgment about the quality of the presented answers.

Based on this, which answer are you more confident in that it answers the question correctly? 1 or 2?
\end{quote}

Let $\mathcal{D} = \{(i, j)\}$ denote the set of pairwise comparisons where model $i$ was preferred over model $j$. The Bradley-Terry model expresses the probability of model $i$ being preferred over model $j$ as:
$$
p_{ij} = \frac{1}{1+e^{\theta_j - \theta_i}}
$$
where $\theta_i$ represents the strength of model $i$. The model strengths are estimated by minimizing the binary cross-entropy loss over the observed comparisons:

$$
\hat{\theta} = \arg\min_{\theta} \sum_{(i,j) \in \mathcal{D}} \text{BCELoss}\left(\frac{1}{1+e^{\theta_j - \theta_i}}, 1\right)
$$
See Table \ref{tab:bt_n_comps} for the number of pairwise comparisons we collected for each benchmark, where $n$ is the number of questions in each benchmark, and $m$ is the number of evaluated models. The full set of pairwise comparisons is $nm(m-1)/2$, which grows quadratically in the number of evaluated models. We  collect the full set of pairwise comparisons only on three out of five benchmarks to limit the cost of evaluation.

\paragraph{Bias correction}
LLM-based pairwise comparisons have been shown to exhibit various types of biases such as length-bias \cite{dubois2024length} and preferring answers from itself \cite{zheng2023judging}. Following \citet{li2024crowdsourced}'s approach, we extend our Bradley-Terry model with additional feature coefficients to control for these biases. Let $\mathcal{D} = \{(i, j, \mathbf{x})\}$ where $\mathbf{x} \in \mathbb{R}^d$ represents context features. For style, we collect the following four features (based on \cite{Li_Angelopoulos_Chiang_2024}): answer length, number of headers, number of bold elements and number of lists. For self-preference \cite{wataoka_self-preference_2025}, we record whether the model giving the answer is from the same model family as the evaluator. The model strengths $\theta$ and feature coefficients $\mathbf{w}$ are estimated by:

$$\hat{\theta}, \hat{\mathbf{w}} = \arg\min_{\theta, \mathbf{w}} \sum_{(i,j,\mathbf{x}) \in \mathcal{D}} \text{BCELoss}\left(\frac{1}{1+e^{\theta_j - \theta_i - \mathbf{w}^\top \mathbf{x}}}, 1\right)$$

\subsubsection{Direct judge (baseline)}
A meaningful baseline to compare against would be to use our model directly as a judge. We will call this the \textit{direct judge} baseline. Instead of collecting pairwise comparisons, we prompt the language model to directly classify freeform answers as either \texttt{correct} / \texttt{incorrect} or \texttt{not\_attempted} based on the question and the provided answer (and nothing else -- no ground truth is provided). See Appendix \ref{app:judge-no-GT} for the exact prompt that was used.

\begin{table}[t]
\centering
\caption{\textbf{Number of pairwise comparisons collected for each benchmark.} $n$ is the number of questions, $m$ is the number of evaluated models. Pairwise comparisons are rounded to the nearest 5,000, with the $nm$ multiplier shown in parentheses. We put an asterisk (*) where we collected the full set of possible pairwise comparisons.}
\label{tab:bt_n_comps}
% \resizebox{\figwidth}{!}{%
\resizebox{\figwidthnarrow}{!}{%
\begin{tabular}{@{}lccc@{}}
\toprule
\textbf{Benchmark} & \textbf{$n$} & \textbf{$m$} & \textbf{\# Comparisons ($\times nm$)} \\ \midrule
MMLU Pro*          & 492          & 12           & $30{,}000$ ($\approx 5.1$)           \\
GPQA Diamond*      & 126          & 10           & $5{,}000$ ($\approx4$)              \\
Simple QA*         & 500          & 11           & $10{,}000$ ($\approx1.8$)\tablefootnote{We filtered out pairs with unattempted responses due to high non-response rates (36-62\% for Claude models).}            \\
GSM8K              & 1318         & 27           & $100{,}000$ ($\approx2.8$)            \\
BBH (Multitask)    & 4583         & 13           & $115{,}000$ ($\approx1.9$)            \\ \bottomrule
\end{tabular}%
}
\end{table}

\section{Experimental Setup}

\begin{figure*}[ht]
    \begin{subfigure}{0.24\linewidth}
        \includegraphics[width=\linewidth]{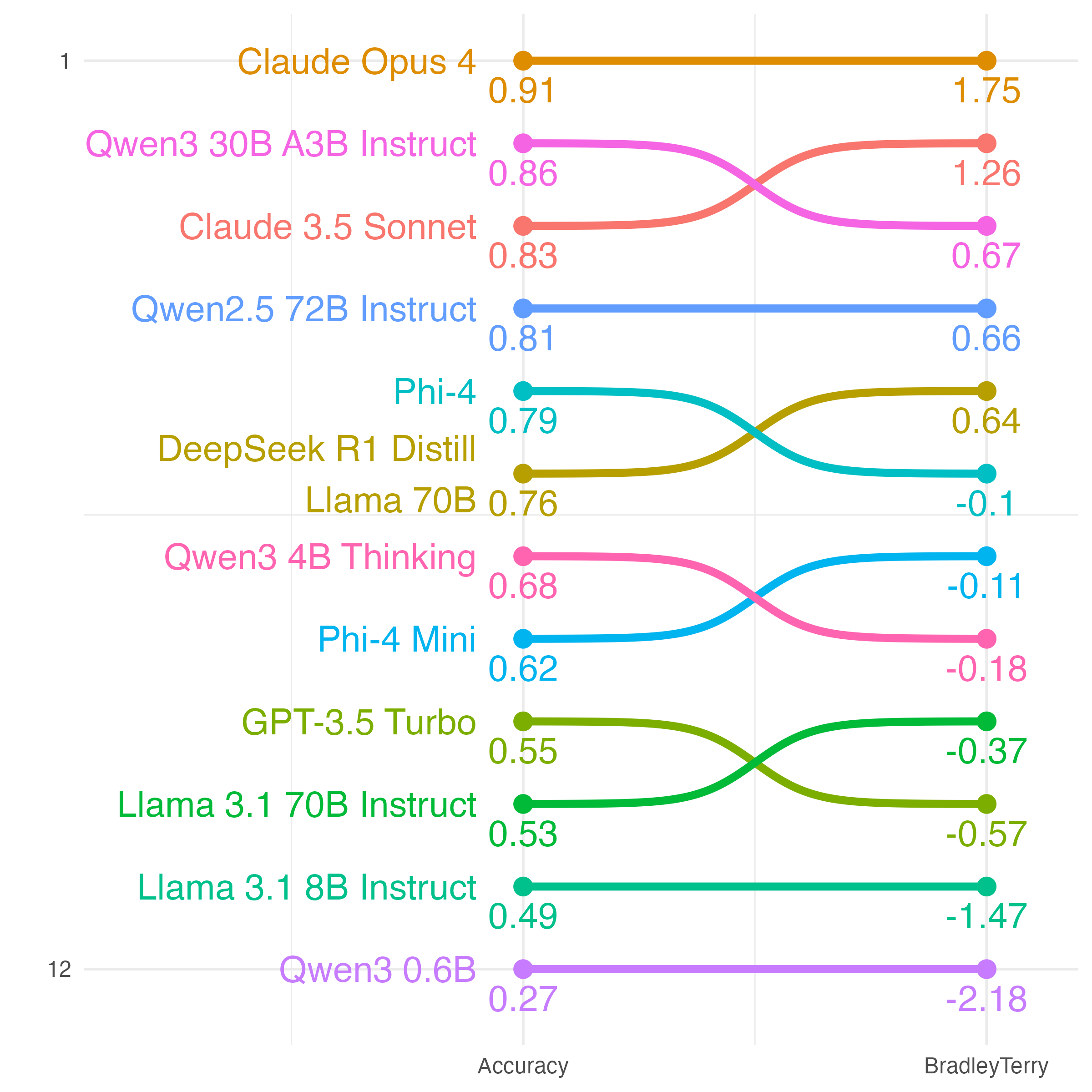}
        \captionsetup{margin={1.5cm,0cm}}
        \caption{MMLU Pro\\$K_D=6\%$}
        \label{subfig:bump_mmlu}
    \end{subfigure}
    \begin{subfigure}{0.24\linewidth}
        \includegraphics[width=\linewidth]{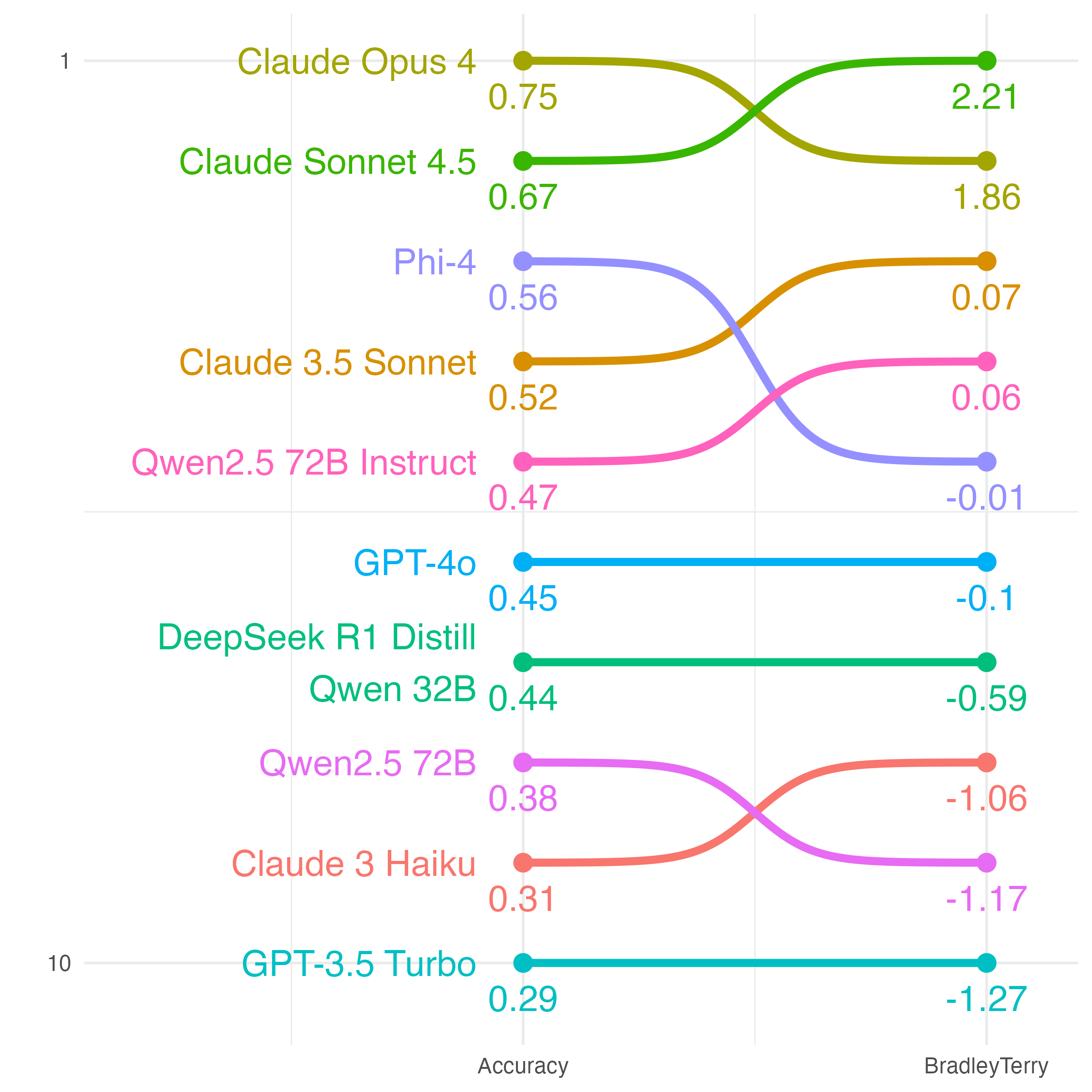}
        \captionsetup{margin={1.5cm,0cm}}
        \caption{GPQA-D\\$K_D=8.9\%$}
        \label{fig:gpqa_bump}
    \end{subfigure}
    \begin{subfigure}{0.24\linewidth}
        \includegraphics[width=\linewidth]{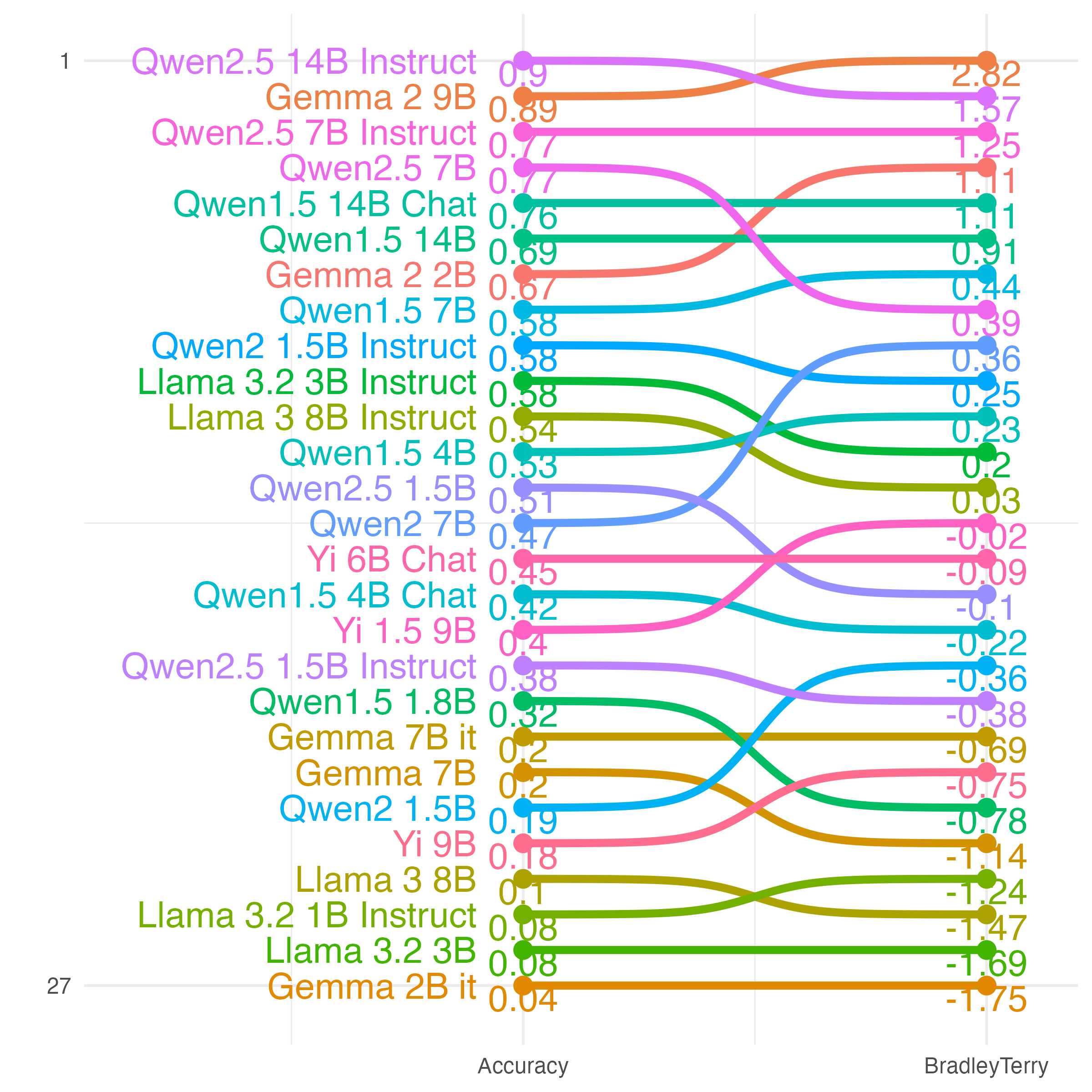}
        \captionsetup{margin={1.5cm,0cm}}
        \caption{GSM8K\\$K_D=7.4\%$}
        \label{fig:gsm8k_bump}
    \end{subfigure}
    \begin{subfigure}{0.24\linewidth}
        \includegraphics[width=\linewidth]{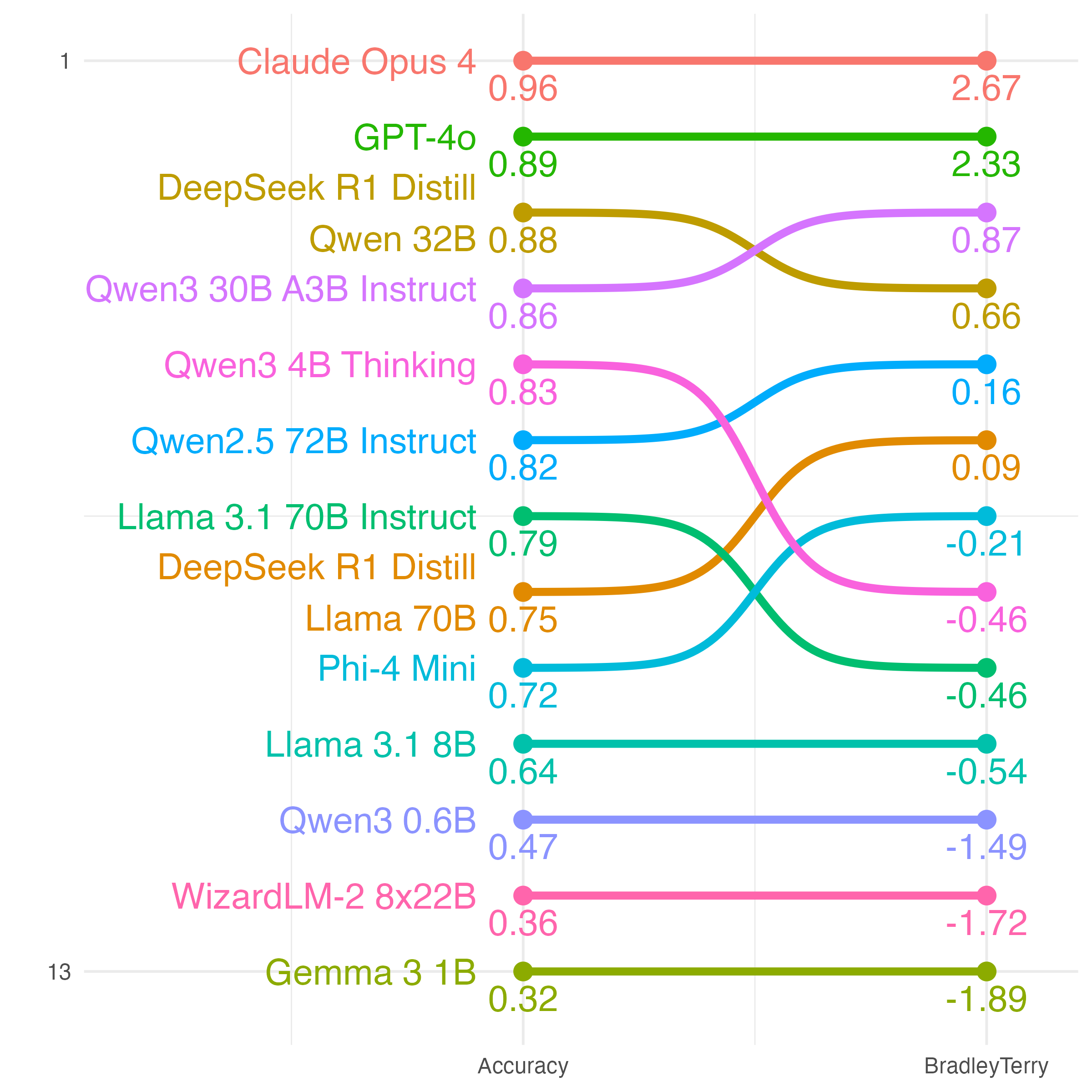}
        \captionsetup{margin={1.5cm,0cm}}
        \caption{BBH\\$K_D=7.7\%$}
        \label{subfig:bump_bbh}
    \end{subfigure}
    \caption{\textbf{Alignment with accuracy based ranking}. On the left side we have models ranked by their accuracy while on the right we rank them by their Bradley-Terry score. Distance between ranks is measured using Kendall's distance ($K_D$). We see consistent alignment across different types of benchmarks: multiple choice (MMLU Pro and GPQA Diamond [GPQA-D]), LLM graded (GSM8K) and multitask (Big Bench Hard [BBH]). Judge model: \texttt{gpt-oss-20b}.}
    \label{fig:bump_plots}
\end{figure*}

\begin{figure}[ht]
    \centering
    \begin{subfigure}{0.49\figwidthnarrow}
        \includegraphics[width=\linewidth]{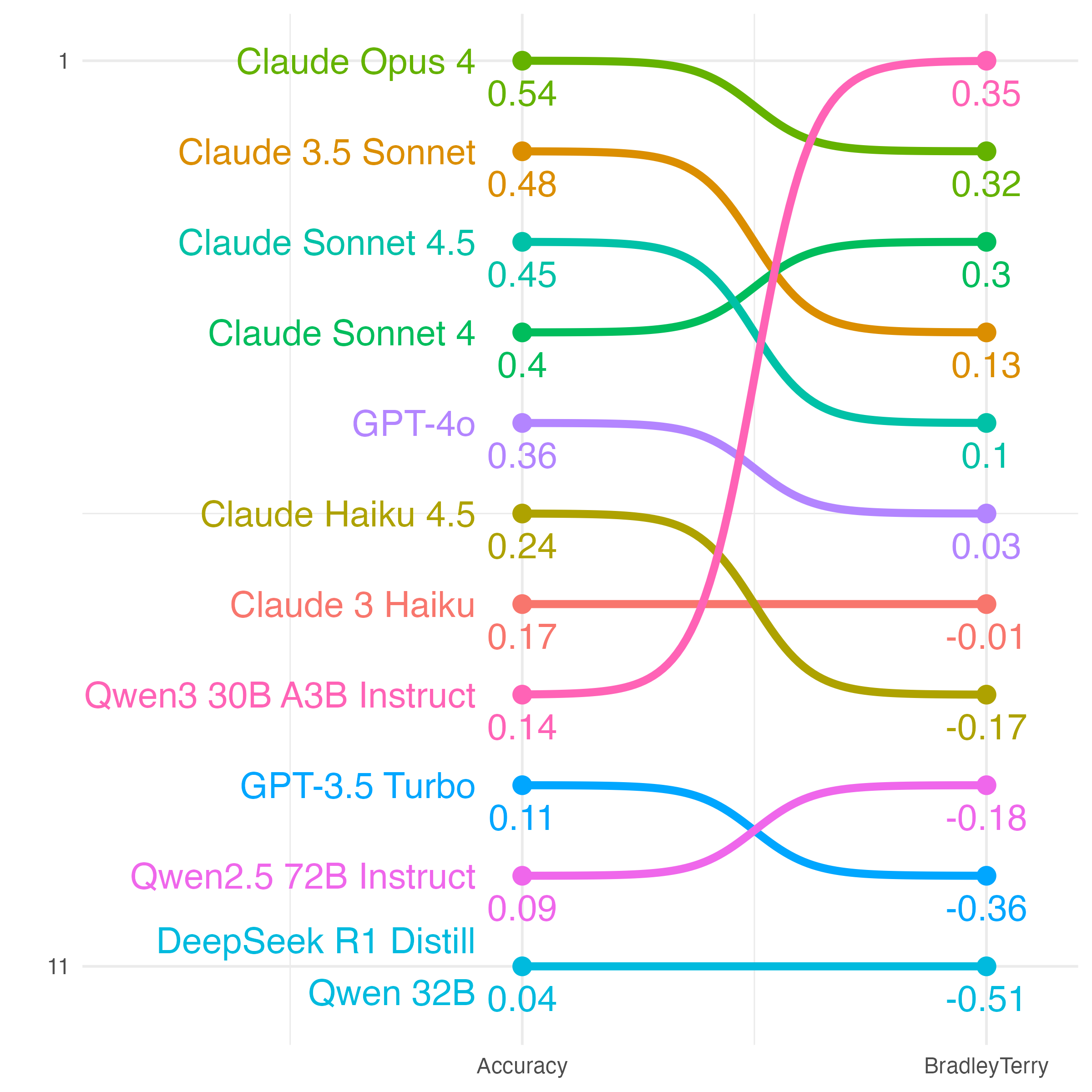}
        \captionsetup{margin={1.5cm,0cm}}
        \caption{Bradley-Terry\\$K_D=20\%$}
        \label{subfig:bump_simpleqa_bt}
    \end{subfigure}
    \begin{subfigure}{0.49\figwidthnarrow}
        \includegraphics[width=\linewidth]{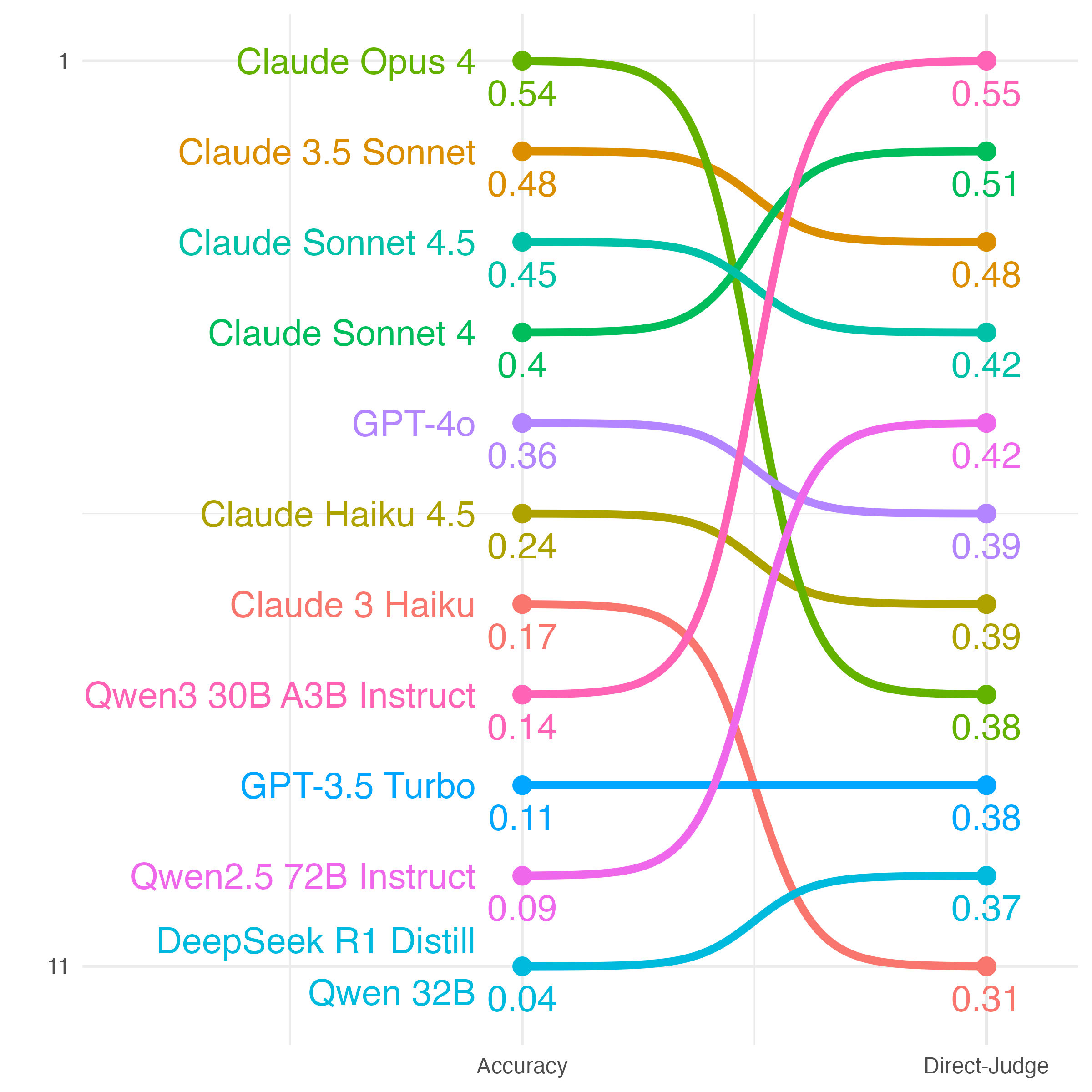}
        \captionsetup{margin={1.5cm,0cm}}
        \caption{Direct Judge\\$K_D=38.2\%$}
        \label{subfig:bump_simpleqa_baseline}
    \end{subfigure}
    \caption{\textbf{Direct judge doubles the rank distance on SimpleQA} We plot the rank alignment on SimpleQA for both Bradley-Terry and the direct judge baseline. Using \texttt{gpt-oss-20b} as a judge directly doubles the model pairs that would need to be swapped to recover the accuracy based ranking. }
    \label{fig:double_distance}
\end{figure}

\subsection{Models}
\paragraph{Evaluated models}
Depending on the benchmark, we ranked between 10-27 models. 
The evaluated models span a range of architectures, training methods and sizes, including open-source and closed-source models. 
For each benchmark we start with a large initial set of models (see Table \ref{tab:models-full} for the full list), which we then filter based on the following criteria:
We keep models that have i) $>90$\% answer parse rate ii) above-baseline accuracy and iii) a difference of at least $1\%$ accuracy between model pairs. This last point ensures that differences in ranking reflect meaningful differences in accuracy.  
See Appendix~\ref{app:filtering} for more details on the filtering criteria and an ablation showing that filtering has only a limited impact on rank correlation results.

\paragraph{Judge models}
We employed three models as judges: OpenAI's open-source \texttt{gpt-oss-120b} and \texttt{gpt-oss-20b}, as well as their \texttt{o3} reasoning model. These models were used in two configurations: (1) collecting pairwise comparisons between pairs of answers without ground truth access  and (2) classifying individual answers as \texttt{correct} / \texttt{incorrect} / \texttt{not\_attempted}. To test generalization beyond OpenAI models, we additionally evaluate two non-reasoning models from different developers: \texttt{phi-4} (Microsoft) and \texttt{gemma-3-27b-it} (Google). Full results for these judges are reported in Appendix~\ref{app:other_models}. 

\subsection{Benchmarks}
We consider five popular benchmarks that cover a range of tasks and difficulty levels: \textbf{MMLU-Pro} \cite{wang2024mmlu}, \textbf{GPQA Diamond} \cite{rein2024gpqa} - two challenging multiple-choice benchmarks. As well as \textbf{GSM8K} \cite{cobbe2021training}, a grade school math benchmark and the more recent \textbf{SimpleQA} \cite{wei2024measuring}. We also include a multitask benchmark: \textbf{Big Bench Hard (BBH)} \cite{suzgun2022h}. MMLU-Pro covers challenging questions across multiple domains, GPQA Diamond focuses on graduate-school level problems, SimpleQA tests open-domain question answering, while BBH offers a diverse suite of challenging reasoning tasks. 

\paragraph{Freeform format}
As previously mentioned, multiple choice questions were converted into freeform by removing the answer options.  
However, not all multiple choice questions have a freeform equivalent - for example tasks where the correct answer depends on the answer options, such as selecting the ``odd-one-out''. We rely on \citet{chandak2025answer}'s freeform versions of the MMLU Pro and GPQA Diamond that select for multiple choice questions that make sense in the freeform format. Additionally, we filter out 5 multiple choice tasks from BBH for similar reasons, leaving us with 17 tasks in total (7 multiple choice and 10 freeform).

\subsection{Metrics}
\paragraph{Score correlation}
We use \textit{Pearson's R} to measure the linear correlation between scores. It captures the strength and direction of the linear relationship.

\paragraph{Rank correlation}
We use \textit{Kendall's Tau} $\tau = \frac{C-D}{C+D}$, where $C$ and $D$ are concordant and discordant pairs as our primary rank correlation measure, and additionally report \textit{Spearman's Rho}, a measure of rank monotonicity. A correlation of 1 means perfect alignment, while a correlation of -1 means the ranks are reversed.

\paragraph{Rank distance}
\textit{Kendall's distance} $K_D = \frac{1-\tau}{2}$ is the fraction of model pairs that must be swapped to transform one ranking into the other.

\paragraph{Interpreting rank correlation}
There is no consensus on what constitutes ``high'' rank alignment. Some prior work uses fixed thresholds of 0.7--0.8 \cite{liu2023question, sun2023validity}, while \citet{perlitz2024these} argue for comparative baselines over absolute thresholds. We report both perspectives: values above 0.9 for Spearman's Rho or Pearson's R would be considered high under fixed-threshold standards, and we compare Bradley-Terry against a direct judge baseline in line with the comparative approach. We prefer Kendall's Tau ($\tau$) for its direct conversion to Kendall's distance ($K_D$) for interpretability.  $\tau>0$ means we have better than random alignment because less than half of the model pairs need to be swapped ($K_D<0.5$).\footnote{We considered defining explicit thresholds for $0<\tau\leq1$ to characterize certain values as ``fair'', ``good'' or ``excellent'', but ultimately decided against them. Even Cohen, who is widely cited for providing such frames of reference, urges the reader ``to avoid the use of these conventions, if he can'' \cite{cohen2013statistical}. }
Our main results are above this random baseline and on 4 out of 5 benchmarks require only 6-9\% of model pairs to be swapped, with Spearman's Rho $>0.9$.

\section{Results}

\subsection{High alignment with accuracy}
We used OpenAI's open-source \texttt{gpt-oss-20b} model to collect pairwise preferences over answer pairs, which were aggregated using Bradley-Terry. Figure~\ref{fig:main} presents the main result of this paper. Each point in the figure represents a model evaluated on one of the benchmarks. We plot the models' accuracy on the x axis, and their Bradley-Terry score on the y axis. What we find is that the Bradley-Terry score has high correlation with accuracy. We observe both high rank correlation (Spearman's Rho $> 0.9$, or Kendall's Tau $>0.8$) as well as fairly high linear correlation (Pearson's R $ > 0.87$). 
This high correlation with accuracy holds true even when we don't collect the full set of pairwise comparisons. For GSM8K and BBH we only collect a fraction ($\sim20-30\%$) of the full pairwise comparisons and obtain as impressive results as in cases where we collected the full set.
We find these results quite encouraging, because they indicate that pairwise comparisons \textit{can} be used to construct rankings similar to the ones one would get from model accuracies, at least in this discriminative setting. While Bradley-Terry scores do not recover exact accuracy values, their strong correlation makes relative differences between models interpretable. Additionally, a common concern with pairwise comparison based evaluation is scalability -- however, we show that even a fraction of the full set of pairs can give us competitive results. Per-dataset rank correlations on the full set of judges are provided in the Appendix (Table~\ref{tab:per_dataset_rank_corr}). In general, more capable judges yield even higher correlations.

Beyond correlation metrics, we examine the distance between rankings using Kendall's distance, which provides an intuitive measure: the fraction of model pairs that must be swapped to transform one ranking into another.
For four out of five benchmarks, we find strong rank alignment: only $6-8.9\%$ of model pairs need to be swapped to recover the accuracy based ranking. Figure~\ref{fig:bump_plots} illustrates this alignment by plotting accuracy based and Bradley-Terry rankings side by side. We observe strong alignment on benchmarks where the judge model does relatively well: \texttt{gpt-oss-20b} ranks in the top 50th percentile on MMLU Pro, GPQA Diamond, GSM8K and BBH. On SimpleQA, however, \texttt{gpt-oss-20b} is the worst-performing model. Despite this, it is able to achieve non-trivial rank alignment of $K_D=20\%$. See Figure \ref{subfig:bump_simpleqa_bt} for a similar comparison on SimpleQA.

\subsection{Comparison to baseline}
Next, we investigate performance when using an LLM as a direct judge rather than collecting pairwise preferences. The direct judge must evaluate each answer independently, requiring $nm$ model calls (one per model-question pair). In comparison, we collected between $2\text{--}5 \times nm$ pairwise comparisons (see Table \ref{tab:bt_n_comps}), which is only a constant factor difference. In Table \ref{tab:bt_vs_baseline} we provide the average rank and score correlations for both Bradley-Terry and the direct judge baseline. On average, the two methods perform comparably. Bradley-Terry achieves slightly higher rank and score correlation for \texttt{gpt-oss-20b}, while both methods yield nearly identical results for \texttt{gpt-oss-120b}. With \texttt{o3} as the evaluator, the direct judge has a slight advantage, though differences remain minor overall. Both methods achieve high rank and score correlation with the accuracy based ranking. In the direct judge setting, this should come as no surprise. If a judge does well on the task it is supposed to evaluate, it should also do well in classifying answers as either correct or incorrect. This is somewhat less straightforward for pairwise comparisons, since many of the answer pairs are non-discriminative. There, the decision could rely on more subjective factors, potentially introducing additional noise.

\begin{figure*}[]
    \centering
    \begin{subfigure}{0.49\linewidth}
        \centering
        \includegraphics[width=0.9\linewidth]{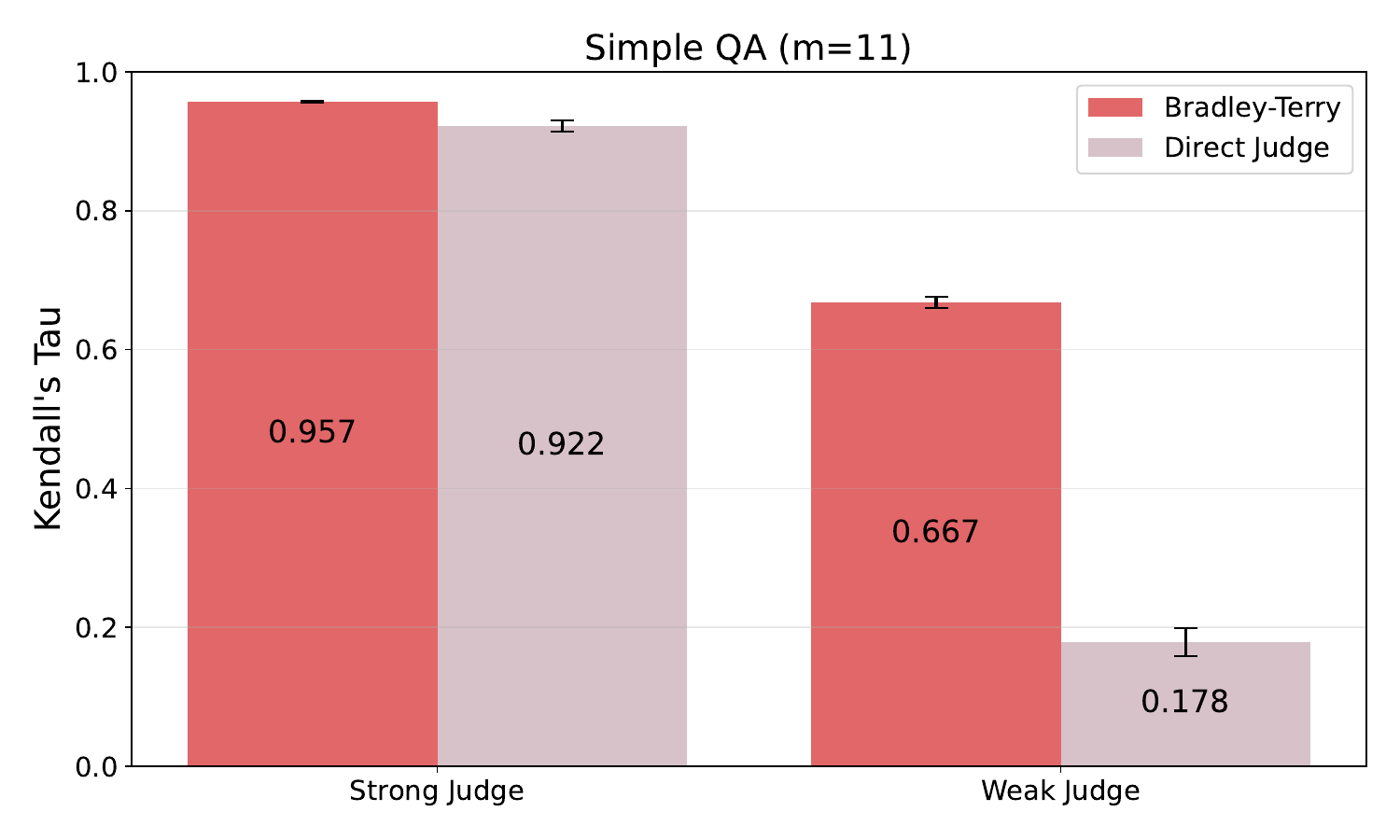}
        \caption{Rank correlation}
    \end{subfigure}
    \hfill
    \begin{subfigure}{0.49\linewidth}
        \centering
        \includegraphics[width=0.9\linewidth]{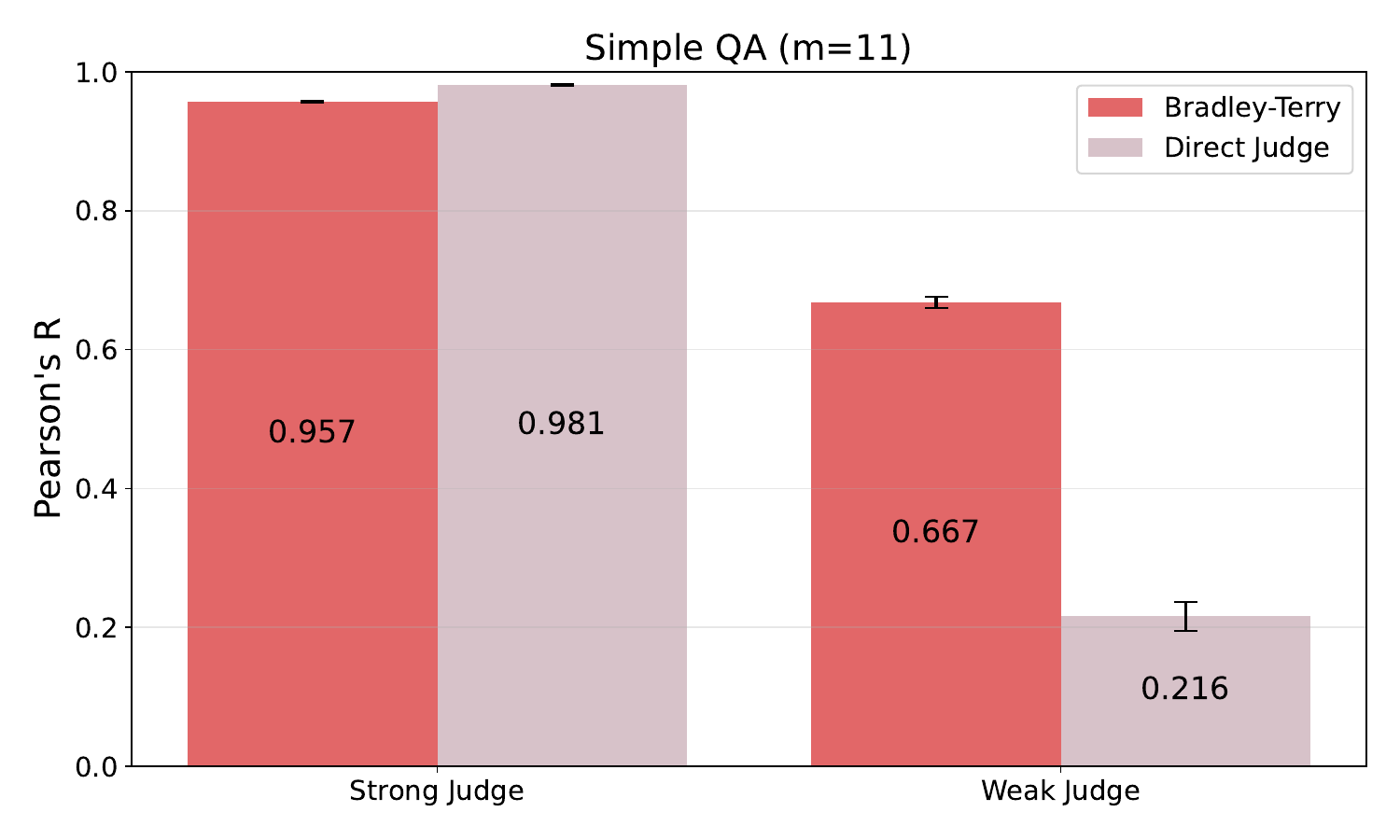}
        \caption{Score correlation}
    \end{subfigure}
    \caption{\textbf{Bradley-Terry significantly outperforms baseline in the weak model setting.} We measure the correlation to the accuracy-based ranking on SimpleQA. We vary our evaluator between the ``\textit{strongest model}''  on the benchmark (\texttt{o3}, 59.3\% acc) and the ``\textit{weakest model}'' on the benchmark (\texttt{gpt-oss-20b}, 4.9\% acc). While the direct judge is better in the strong model setting, Bradley-Terry clearly dominates the baseline in the weak model setting. Error bars are 95\% confidence intervals on 100 bootstrap samples.}
    \label{fig:bt_vs_baseline_simpleqa}
\end{figure*}

While both methods demonstrate strong alignment with accuracy-based rankings, their practical utility is most relevant in \textit{frontier evaluation settings}, where ground truth is often unavailable and we cannot determine a priori which judge will perform best. To illustrate this, we examine the best and worst performing models on our most challenging benchmark, SimpleQA. Figure \ref{fig:bt_vs_baseline_simpleqa} shows that if we use the top-performing judge on the task (\texttt{o3} with 59.3\% accuracy), direct LLM-as-a-judge evaluation and Bradley-Terry are comparable. However, if we use the weakest judge (\texttt{gpt-oss-20b}, 4.9\% accuracy) -- which would be indistinguishable from stronger models in the absence of ground truth -- the results change substantially.
In the weak model setting, Bradley-Terry substantially outperforms the direct judge approach. In fact, the direct judge almost doubles the rank distance compared to Bradley-Terry: from $20\%$ to $38.2\%$ (see Figure \ref{fig:double_distance}).
This suggests Bradley-Terry provides more stable rankings when evaluator quality is uncertain, though direct evaluation may perform better with well-selected judges. We find this pattern holds across 3 out of 4 weak judges (on both \texttt{gpt-oss} models and \texttt{phi-4}). The one exception is \texttt{gemma-3-27b-it} where the direct judge performs better because its errors happen to correlate with the accuracy-based ranking. See Appendix~\ref{app:weak_judge} for the full discussion.

\begin{table}[]
\centering
\caption{\textbf{On average, Bradley-Terry achieves comparable results to the Direct Judge baseline}. We report the mean and standard deviation across benchmarks for rank and score correlation metrics.}
\label{tab:bt_vs_baseline}
\resizebox{\figwidth}{!}{%
\begin{tabular}{@{}lcccc@{}}
\toprule
 & \multicolumn{2}{c}{\textbf{Rank correlation ($\tau$)}} & \multicolumn{2}{c}{\textbf{Score correlation}} \\ 
 & Bradley-Terry          & Direct Judge         & Bradley-Terry          & Direct Judge          \\ \midrule
\texttt{gpt-oss-20b}  & $0.80 \pm 0.10$ & $0.76 \pm 0.26$ & $0.88 \pm 0.10$ & $0.79 \pm 0.29$ \\
\texttt{gpt-oss-120b} & $0.82 \pm 0.07$ & $0.83 \pm 0.09$ & $0.91 \pm 0.05$ & $0.88 \pm 0.10$ \\
\texttt{o3}           & $0.81 \pm 0.05$ & $0.85 \pm 0.09$ & $0.93 \pm 0.03$ & $0.96 \pm 0.02$ \\ \bottomrule
\end{tabular}%
}
\end{table}

\begin{figure}
    \centering
    \includegraphics[width=\figwidthnarrow]{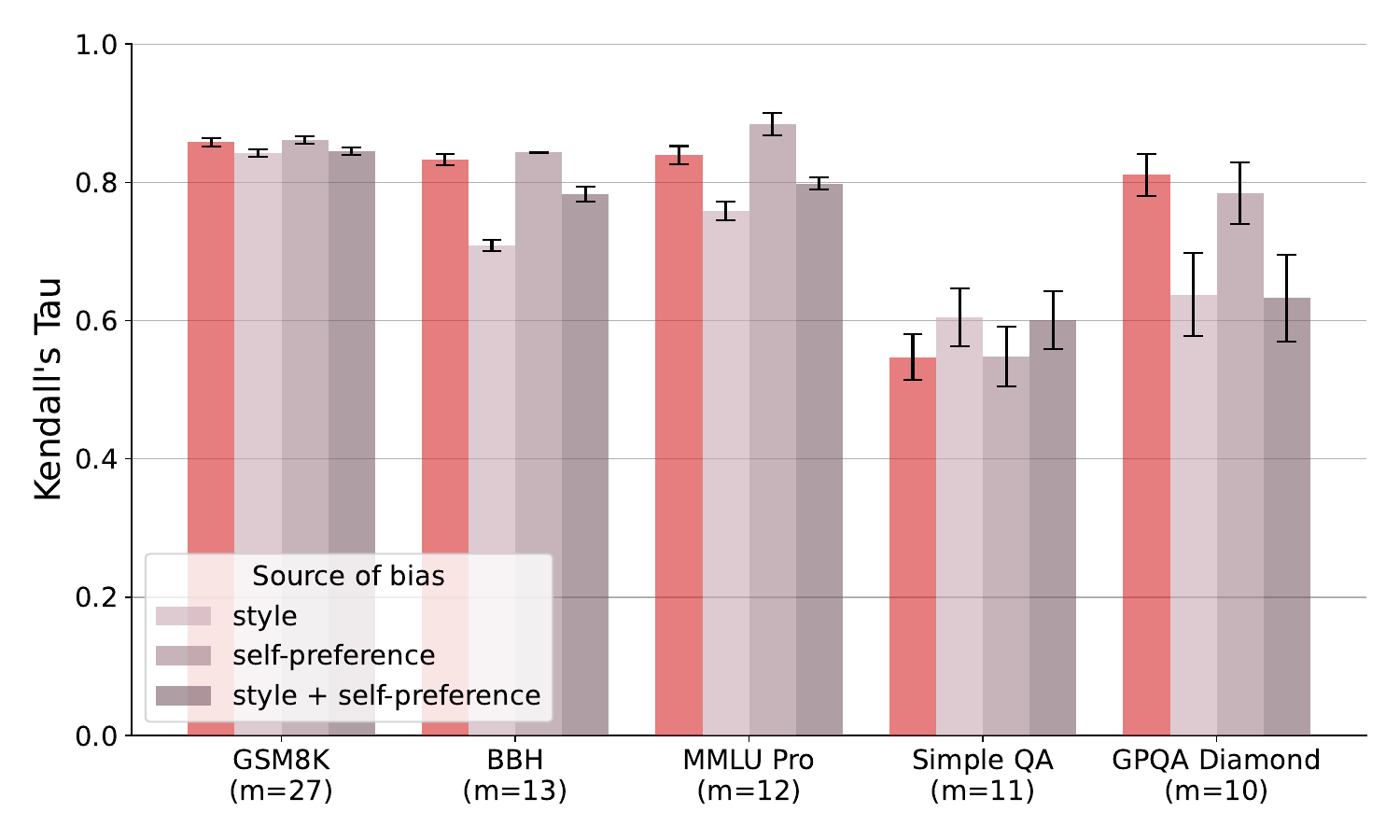}
    \caption{\textbf{Correcting for biases can lead to modest improvements} Original rank correlation (without bias correction) is \textcolor{customred}{red}. $m$ stands for number of ranked models. Error bars are 95\% confidence intervals on 100 bootstrap samples. Judge model: \texttt{gpt-oss-20b}}
    \label{fig:bias}
\end{figure}

\subsection{Bias correction}

We also investigate potential sources of judge bias documented in prior work \cite{zheng2023judging, dubois2024length}. In Figure \ref{fig:bias}, we correct for three types of biases: style, self-preference, and a combination of both. \textit{Style} controls for judges' preferences for answer length and formatting. \textit{Self-preference} controls for preference for answers from the same model family as the judge (OpenAI models in our case). We study whether correcting for such biases leads to improvements in rank correlation compared to our original results. We find that bias correction offers modest improvements on some of our benchmarks. For \texttt{gpt-oss-20b}, we see some improvement in rank correlation on MMLU Pro. Bias correction on other judges shows similar, minor improvements in rank correlation (see Appendix \ref{app:bias}). These findings do not imply the absence of bias, but rather that such biases have limited impact on model ranking in our evaluation. This is perhaps surprising: almost $58\%$ of our pairwise comparisons are \textit{non-discriminative} — pairs where both answers are correct or both are incorrect — meaning the judge cannot rely on correctness and must instead fall back on other signals, such as appearance. Yet correcting for these stylistic signals barely moves the ranking — even though they are the only available cue in many cases. As we show in the next section, non-discriminative pairs are not uninformative. On the contrary, some of the appearance-based cues judges fall back on turn out to be correlated with model correctness, carrying genuine predictive signal.

\subsection{Signal in non-discriminative pairs}

\begin{figure*}[t]
    \centering
    \begin{subfigure}{0.49\textwidth}
        \centering
        \includegraphics[width=\linewidth]{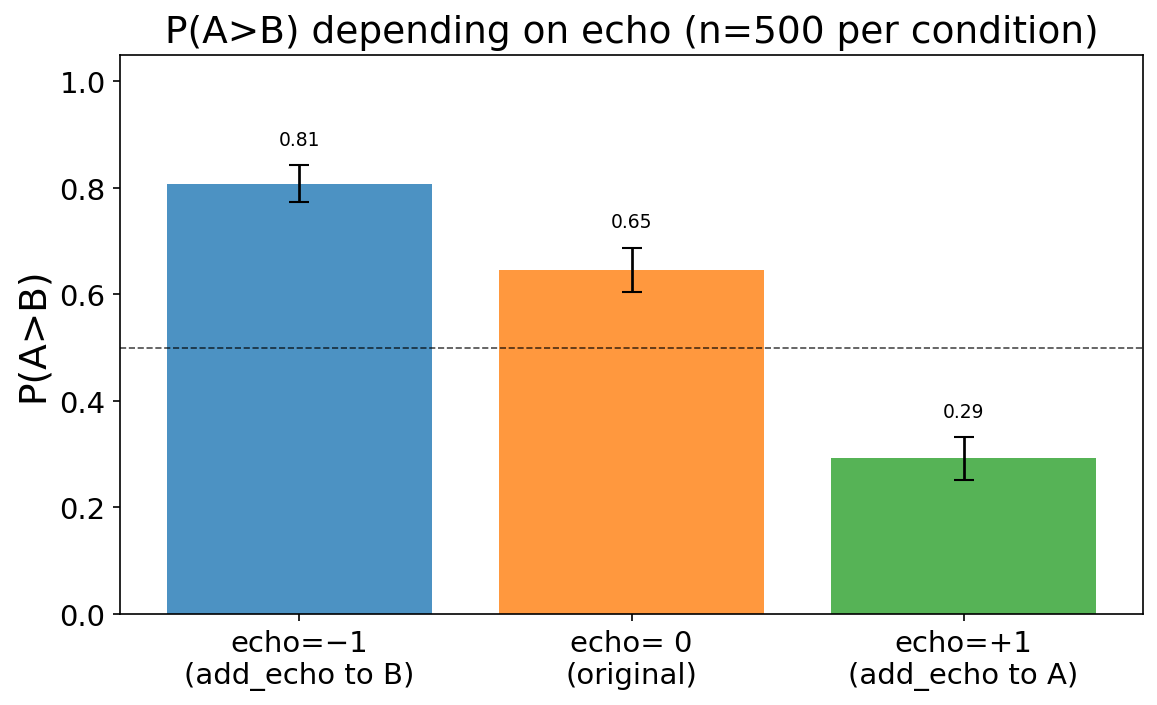}
        \caption{Echo has a significant effect on judge preference.}
        \label{fig:echo_effect}
    \end{subfigure}
    \hfill
    \begin{subfigure}{0.49\textwidth}
        \centering
        \includegraphics[width=\linewidth]{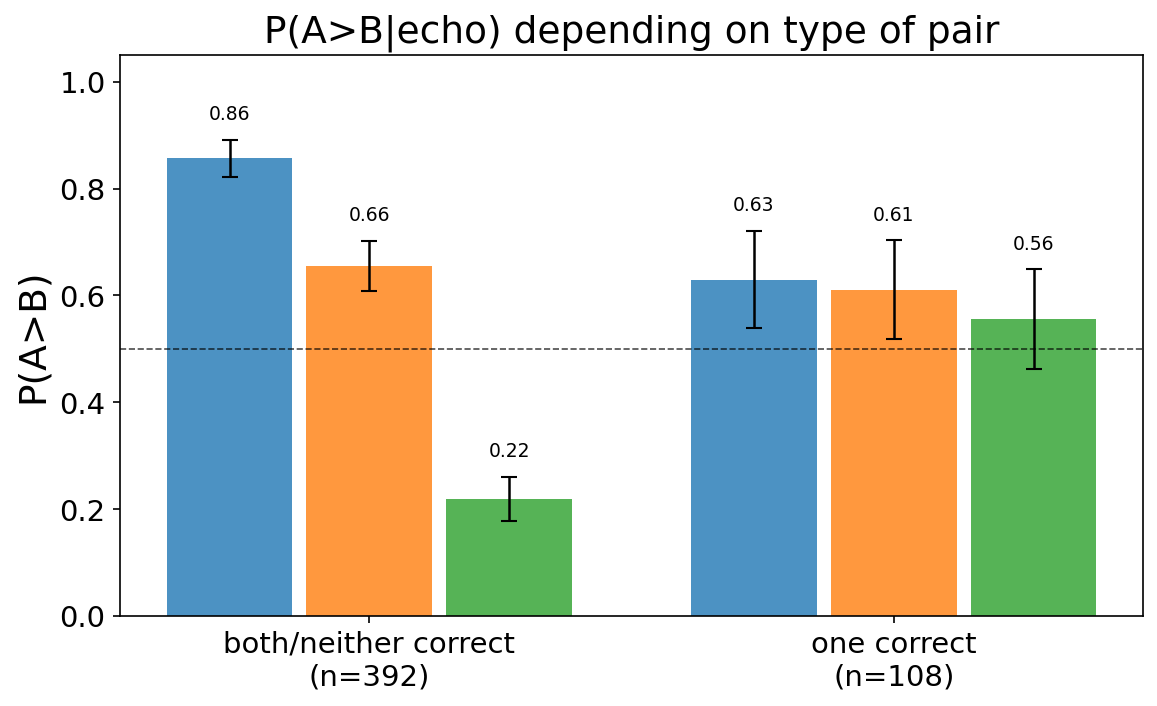}
        \caption{Echo matters only on non-discriminative pairs.}
        \label{fig:echo_effect_stratified}
    \end{subfigure}
    \caption{\textbf{Echo as a causal driver of judge preference.} We plot the probability of the judge model choosing answer A over B. The intervention is adding echo to one answer by appending the question--answer sequence three times. Echo is a strong driver of judge preference on non-discriminative pairs (where both answers are either correct or incorrect). This effect disappears on discriminative pairs. Judge model: \texttt{o3}. Error bars are 95\% confidence intervals.}
    \label{fig:echo_effect_combined}
\end{figure*}
Non-discriminative pairs (where both answers are either correct or incorrect) contain a surprising amount of useful information. Compared to using all available pairs, Kendall's Tau decreases from 0.8 to 0.68 ($K_D=16\%$), while Pearson's R drops from 0.88 to 0.8 (Fig. \ref{fig:cc_ii}). In the complementary experiment, retaining only \textit{discriminative} pairs (one answer correct, the other incorrect) — which make up $26-37\%$ of total pairs depending on the benchmark — yields the strongest signal (Fig. \ref{fig:ic_ci}): on three out of five benchmarks, correlation is similar to or higher than the full set. On GSM8K, Kendall's distance drops from $7.4\%$ to merely $3.4\%$, and on BBH from $7.7\%$ to $3.8\%$. On the remaining two benchmarks (SimpleQA with 3,061 pairs and GPQA Diamond with 1,956 pairs), we suspect the sample size was too small. Results on non-discriminative and discriminative pairs can be found in Appendix \ref{app:filtered_pairs}.

What drives the signal in non-discriminative pairs? We identify \textit{echo} as a causal driver (Figure~\ref{fig:echo_effect_combined}). Echo is a failure mode where a model fails to stop after its final answer and repeats content, such as duplicating phrases or re-generating the original question-answer template. Figure~\ref{fig:pipeline} shows an example of such behavior: the judge prefers Claude's answer over Qwen's, despite both models giving correct answers. In this example, Qwen repeatedly duplicates the final sentence of its response, likely influencing the judge's preference for Claude. Echo is strongly correlated with both judge preference and answer correctness (Spearman $\rho \approx -0.50$). To test causality, we ran a controlled intervention on 500 BBH pairs: adding echo to answer A drops $P(A>B)$ from 0.65 to 0.29, while adding it to B raises it to 0.81 (contrast $-0.52$, 95\% CI $[-0.57, -0.46]$, see Figure~\ref{fig:echo_effect}). Crucially, this effect disappears on discriminative pairs (contrast $-0.07$, 95\% CI $[-0.21, +0.06]$, see Figure~\ref{fig:echo_effect_stratified}) — when one answer is objectively correct, the judge relies on correctness rather than echo. We detected echo in 65\% of BBH answers; in 46.7\% of non-discriminative pairs exactly one answer is flagged, making echo a potential driver in nearly half of those comparisons. Echo avoidance thus acts as a quality cue which causally drives judge preference and accounts for a meaningful portion of the signal on non-discriminative pairs. See Appendix~\ref{app:echo} for details on how we detect echo.

\section{Limitations}
\paragraph{Discriminative tasks only.}
Our results are scoped to benchmarks with ground-truth answers, which is what enables a controlled comparison with accuracy-based rankings. Whether pairwise rankings reflect answer correctness in open-ended settings — where ground truth is unavailable — remains an open question.

\paragraph{MCQ and freeform rankings may differ.}
We convert multiple-choice benchmarks to freeform because MCQ answers are typically single letters, making it impossible to study whether judge preferences are driven by stylistic cues or correctness. Freeform also eliminates MCQ shortcuts such as selecting the odd-one-out option \cite{balepur2025these}. However, this means we compare a freeform Bradley-Terry ranking against an MCQ accuracy ranking. If the two formats produce different model orderings, our reported rank correlation is a conservative underestimate of alignment within either setting.

% \paragraph{No standard for ``high'' rank correlation.}
% There is no consensus threshold in benchmarking: fixed thresholds around 0.7-0.8 have been used \cite{liu2023question, sun2023validity}. Our main results show Spearman's $\rho > 0.9$, which should be considered high by most standards. In contrast, \citet{perlitz2024these} recommend comparative approaches, at least 10 models, and finer granularities. We satisfy the first two: comparing against a Direct Judge baseline across 10-27 models. Aggregate metrics can also mask individual misplacements that violate intuitive alignment — on SimpleQA, the 8th-ranked model is placed first, yet Kendall's distance still reads as ``fair'' ($K_D=20\%$), because it averages over all pairwise swaps. Reporting at finer granularities — e.g., alignment among the top-$k$ models — would surface such cases and remains future work.

\paragraph{Rank correlation may not reflect intuitive alignment}
We note that the reported rank correlation metrics do not distinguish large rank displacements from multiple small ones. For example, in a 10-model ranking, ranking the first model as sixth or swapping five adjacent model pairs is equivalent in terms of Kendall's distance. They also treat misalignment at the top and the bottom of the leaderboard equivalently.

\paragraph{Rankings reflect relative, not absolute, performance.}
Bradley-Terry ranks models relative to each other. If all models fail a task, a high Bradley-Terry score still only identifies relative differences. However, it does not tell us if the ''best'' model is actually good at some task. In other words, it does not detect collective incompetence.

\section{Discussion}
We studied whether pairwise comparison based evaluation can be used to produce rankings consistent with accuracy based rankings in a freeform generative evaluation setting.
Our results show that Elo-style evaluation achieves high agreement with the ordering of models one would get in labeled evaluation. Not only do we get high rank alignment, but also the Bradley-Terry score correlates highly with accuracy.
This finding is surprising for multiple reasons. For one, there have been reservations about what pairwise comparisons actually measure. Many have found that surface-level appearance or judge biases could potentially affect model rankings. We show that in our case, correcting for such biases results only in modest improvements. While such biases may affect absolute numbers, their effect on relative model ranking is less pronounced.
Moreover, pairwise comparisons contain many examples where judgment often has to be made based on superficial cues. Nevertheless, even non-discriminative pairs contain useful signal and we identify echo as a causal driver of judge preference on such pairs.
Furthermore, in the weak judge regime, pairwise preference based rankings substantially outperform direct LLM evaluation. This suggests that Elo-style rankings provide a more robust alternative at the evaluation frontier, where the quality of the evaluator is uncertain. This finding holds across 3 out of 4 weak-judge cases (see Appendix \ref{app:weak_judge} for full discussion).

Understanding the full scope of the validity of model rankings from pairwise comparisons remains an important question. With our results we attempt to provide some clarity in cases where controlled comparison with the ground truth based ranking is possible.
Several directions remain open for future work. First, our evaluation does not assess robustness to adversarial gaming, where a model is optimized to win pairwise comparisons rather than to perform well on the task. Second, in the discriminative setting, cheaper alternatives such as self-consistency or confidence-based signals may achieve comparable rankings at lower cost; a direct cost-efficiency comparison would sharpen the practical positioning of pairwise evaluation.
%We note that these findings derive from benchmarks with available ground truth, which enabled our controlled comparison.  

\section*{Acknowledgements}
The authors thank the International Max Planck Research School for Intelligent Systems (IMPRS-IS) for supporting Mina Remeli. This work was supported by the Tübingen AI Center. The authors are grateful to Guanhua Zhang and Nikhil Chandak for providing access to their data, to Florian E. Dorner, Vivian Y. Nastl and Ricardo Dominguez-Olmedo for their feedback on our experiments and to Ana-Andreea Stoica and Mila Gorecki for their help during revisions. We also thank the anonymous reviewers for their valuable feedback.

\section*{Impact Statement}
Our work is a controlled validation study of pairwise comparison-based evaluation methods for ranking large language models on verifiable benchmarks. Pairwise comparisons are widely used to evaluate models on open-ended tasks and increasingly shape leaderboards, model-selection decisions, and post-training signals; our results bear on whether the same methodology can be extended to settings where ground truth is available but expensive to collect. Broader adoption could make evaluation cheaper in terms of labeling cost and also more flexible by virtue of being automated. However, the extent to which our findings generalize beyond verifiable tasks remains unclear. Furthermore, judge biases could be exploited adversarially. 
To address these concerns, follow-up work could extend the characterization of judge decisions, in particular studying when a judge defaults to superficial cues rather than reasoning grounded in the task. Such insights into the underlying mechanisms would help us weigh the risks and benefits of using this method more broadly.

% biblography
\bibliographystyle{plainnat}
\bibliography{refs-validated}

% appendix
\newpage
\appendix
\section{Appendix}
\begin{table}[h]
  \centering
  \caption{Initial set of models for MMLU Pro, GPQA Diamond, SimpleQA and BBH. For GSM8K, we use the 48 models evaluated by \citet{zhang2025train} as our initial set.}
  \label{tab:models-full}
  \begin{tabular}{@{}ll@{}}
    \toprule
    1  & WizardLM-2 8x22B              \\
        2  & GPT-3.5 Turbo                 \\
        3  & GPT-4o                        \\
        4  & o4-mini                       \\
        5  & Claude 3.5 Sonnet             \\
        6  & Claude Opus 4                 \\
        7  & Claude Haiku 4.5              \\
        8  & Claude Sonnet 4               \\
        9  & Claude 3 Haiku                \\
        10 & Claude Sonnet 4.5             \\
        11 & DeepSeek R1 Distill Llama 70B \\
        12 & DeepSeek R1 Distill Qwen 32B  \\
        13 & Gemma 3 1B                    \\
        14 & Phi-4                         \\
        15 & Phi-4 Mini                    \\
        16 & Llama 3.1 8B                  \\
        17 & Llama 3.1 8B Instruct         \\
        18 & Llama 3.1 70B Instruct        \\
        19 & Qwen3 0.6B                    \\
        20 & Qwen3 32B                     \\
        21 & Qwen3 30B A3B Instruct        \\
        22 & Qwen2.5 72B                   \\
        23 & Qwen2.5 72B Instruct          \\
        24 & Qwen3 4B Thinking             \\
    \bottomrule
  \end{tabular}
\end{table}

\section{Filtering evaluated models} \label{app:filtering}
We started with an initial set of models (see Table \ref{tab:models-full}), and filtered them based on three criteria, in the following order:
\begin{enumerate}
  \item \textbf{Answer parse rate $>90\%$}: We only considered models where we could parse the correct answer letter $>90\%$ of the time. For SimpleQA, we required $>90\%$ valid judge answers. This ensures that we have a sufficient number of valid responses to compute accuracy. Models with low parse rates are often not answering the question at all, or have issue with the following the answer format, and hence their accuracy is not meaningful.
  \item \textbf{Above-baseline accuracy}: We only considered models that were above the baseline accuracy (10\% for MMLU-Pro, 25\% for GPQA Diamond, 1\% for SimpleQA and 5\% for GSM8K). This ensures we only include models that have some meaningful ability on the task.
  \item \textbf{1\% accuracy difference between models}: Each selected model pair should have at least 1\% difference in accuracy. This was done by sorting models by accuracy, then for each model pair that was within 1\% accuracy of each other, we removed the i) one with lower parse rate, or alternatively ii) the one with lower accuracy. If both matched, we randomly selected one for removal.
\end{enumerate}

In the first step we dropped between 5-35\% of models across benchmarks (16.2\% on average), in the second step we dropped 0-10\% of models (5.2\% on average), and in the third step we dropped 13-29\% of models (22.2\% on average).

\subsection{Impact on main results}
Out of these three filtering criteria, the last one had the biggest impact on the number of evaluated models, with an average of 22.2\% of models being dropped. To ensure that our results are not artificially inflated by the last filtering step (essentially removing models that are closest in accuracy), we re-ran our main analysis with \texttt{gpt-oss-20b} on the benchmark with the highest percentage of dropped models (GSM8K, 29\%). This increased the number of evaluated models from the original 27 to 41 and the number of pairwise comparisons from 100k to 250k. We found that the rank correlation dropped only slightly (from 0.85 to 0.82) for Bradley-Terry. We observed a similar drop in score correlation for the Direct Judge baseline as well (from 0.91 to 0.88), suggesting a limited impact of the last filtering step on our main results.

\section{Comparing rank aggregation methods} \label{app:rank_agg}
We evaluated four different rank aggregation methods: two offline methods (WinRate and Bradley-Terry) as well as two popular online methods (Elo and its probabilistic extension TrueSkill). We found that Bradley-Terry had the highest rank correlation with accuracy on three out of five benchmarks, and the highest mean rank correlation across benchmarks (see Table \ref{tab:rank_agg_comparison}). Also overall the offline methods have higher rank correlation than the online methods, which is expected given that Elo and Trueskill are designed for \textit{dynamic} settings which are meant to track performance over time. Given the static nature of accuracy-based evaluation we chose Bradley-Terry as our main rank aggregation method.

\begin{table}[ht]
\centering
\caption{Rank correlation (Kendall's Tau) with different rank aggregation methods. Bradley-Terry beats the other methods on three out of five benchmarks, and has the highest mean rank correlation across benchmarks. Judge: \texttt{gpt-oss-20b}.}
\begin{tabular}{lcccc}
\toprule
Dataset & WinRate & Bradley-Terry & Elo & TrueSkill \\
\midrule
MMLU-Pro & 0.8182 & \textbf{0.8788} & 0.7879 & 0.7879 \\
GPQA-Diamond & 0.8222 & 0.8222 & 0.8667 & \textbf{0.9111} \\
SimpleQA & 0.5273 & \textbf{0.5636} & 0.4182 & 0.4909 \\
GSM8K & \textbf{0.8575} & 0.8519 & 0.7265 & 0.8006 \\
BBH & \textbf{0.8462} & \textbf{0.8462} & 0.7949 & \textbf{0.8462} \\
\midrule
Mean & 0.7743 & 0.7925 & 0.7188 & 0.7673 \\
Std & 0.1244 & 0.1159 & 0.1568 & 0.1448 \\
\bottomrule
\end{tabular}
\label{tab:rank_agg_comparison}
\end{table}

\section{Other judges} \label{app:other_models}
We extend our results to two non-OpenAI judges, \texttt{phi-4} and \texttt{gemma-3-27b-it}, confirming that BT continues to work using non-reasoning models (mean $\tau = 0.60$ and $0.65$ respectively, see Table~\ref{tab:per_dataset_rank_corr}).

\begin{table}[ht]
\centering
\caption{Per-dataset rank correlation (Kendall's Tau). Bold indicates the higher of BT and DJ for each judge-dataset pair.}
\begin{tabular}{lcccccccccc}
\toprule
 & \multicolumn{2}{c}{gpt-oss-20b} & \multicolumn{2}{c}{gpt-oss-120b} & \multicolumn{2}{c}{o3} & \multicolumn{2}{c}{phi-4} & \multicolumn{2}{c}{gemma-3-27b-it} \\
\cmidrule(lr){2-3} \cmidrule(lr){4-5} \cmidrule(lr){6-7} \cmidrule(lr){8-9} \cmidrule(lr){10-11}
 & BT & DJ & BT & DJ & BT & DJ & BT & DJ & BT & DJ \\
\midrule
mmlu\_pro & 0.88 & \textbf{0.91} & \textbf{0.88} & 0.85 & 0.73 & \textbf{0.88} & \textbf{0.91} & -0.09 & \textbf{0.88} & 0.67 \\
gpqa\_diamond & 0.82 & \textbf{0.87} & 0.78 & \textbf{0.91} & \textbf{0.82} & 0.69 & \textbf{0.51} & 0.07 & \textbf{0.69} & 0.38 \\
simple\_qa & \textbf{0.56} & 0.24 & \textbf{0.71} & 0.66 & 0.82 & \textbf{0.96} & \textbf{0.16} & -0.48 & 0.24 & \textbf{0.67} \\
gsm8k & 0.85 & \textbf{0.91} & \textbf{0.87} & 0.86 & 0.86 & \textbf{0.89} & \textbf{0.59} & 0.19 & 0.70 & \textbf{0.71} \\
bbh & 0.85 & \textbf{0.90} & \textbf{0.90} & 0.85 & \textbf{0.87} & 0.85 & \textbf{0.85} & 0.44 & \textbf{0.74} & 0.62 \\
\midrule
Mean & 0.79 & 0.76 & 0.83 & 0.83 & 0.82 & 0.85 & 0.60 & 0.03 & 0.65 & 0.61 \\
Std & 0.12 & 0.26 & 0.07 & 0.09 & 0.05 & 0.09 & 0.27 & 0.30 & 0.22 & 0.12 \\
\bottomrule
\end{tabular}

\label{tab:per_dataset_rank_corr}
\end{table}

\subsection{Weak judge regime on SimpleQA} \label{app:weak_judge}
 Across the four weak-judge cases in our evaluation (\texttt{gpt-oss-20b} 4.9\%, \texttt{gpt-oss-120b} 8.8\%, \texttt{phi-4} 4.5\%, and \texttt{gemma-3-27b-it} 9.6\% accuracy on SimpleQA), BT outperforms DJ in 3 out of 4, supporting our claim that BT is generally more robust in the weak-judge regime. On SimpleQA in particular, the DJ baseline can fail catastrophically ($\tau = -0.48$ with \texttt{phi-4}) or perform surprisingly well ($\tau = 0.67$ with \texttt{gemma-3-27b-it}) for reasons unrelated to judge capability. We examine each case below.

\textbf{\texttt{phi-4}} as a direct judge on SimpleQA is unreliable ($\tau = -0.48$) compared to the Bradley-Terry approach ($\tau = 0.16$). The collapse of \texttt{phi-4} as a direct judge is explained by a systematic misclassification: \texttt{phi-4} labels 76.7\% of ground-truth \texttt{not\_attempted} responses as \texttt{incorrect} (1,547 out of 2,018), correctly identifying only 3.7\% as \texttt{not\_attempted}. This ranks models with well-calibrated uncertainty (such as the Claude models) lower, resulting in the negative correlation with \textit{correct given attempted} accuracy on SimpleQA.

\textbf{\texttt{gemma-3-27b-it}} shows the opposite pattern on SimpleQA — BT worse than DJ (BT $\tau=0.24$ versus DJ $\tau=0.67$). Importantly, this is not because \texttt{gemma-3-27b-it} is a good direct judge — across 5,500 judgments, it marks 91.3\% of attempted responses as correct, leaving only 294 incorrect classifications in total. DJ works here because these sparse incorrect labels happen to correlate with true model accuracy: the three lowest-accuracy models (4.4–10.9\% true accuracy) receive 7.6–21.6\% incorrect labels from \texttt{gemma-3-27b-it}, while the three highest-accuracy models (44.6–53.8\% true accuracy) receive only 1.2–2.2\% — an order of magnitude difference. In other words, \texttt{gemma-3-27b-it}'s errors are not random; they align with the true ranking, which is sufficient for DJ to recover a good ordering.

\subsection{High correlation with accuracy} \label{app:corr_other_models}

\begin{figure}[H]
    \centering
    \begin{subfigure}{0.49\linewidth}
        \centering
        \includegraphics[width=\linewidth]{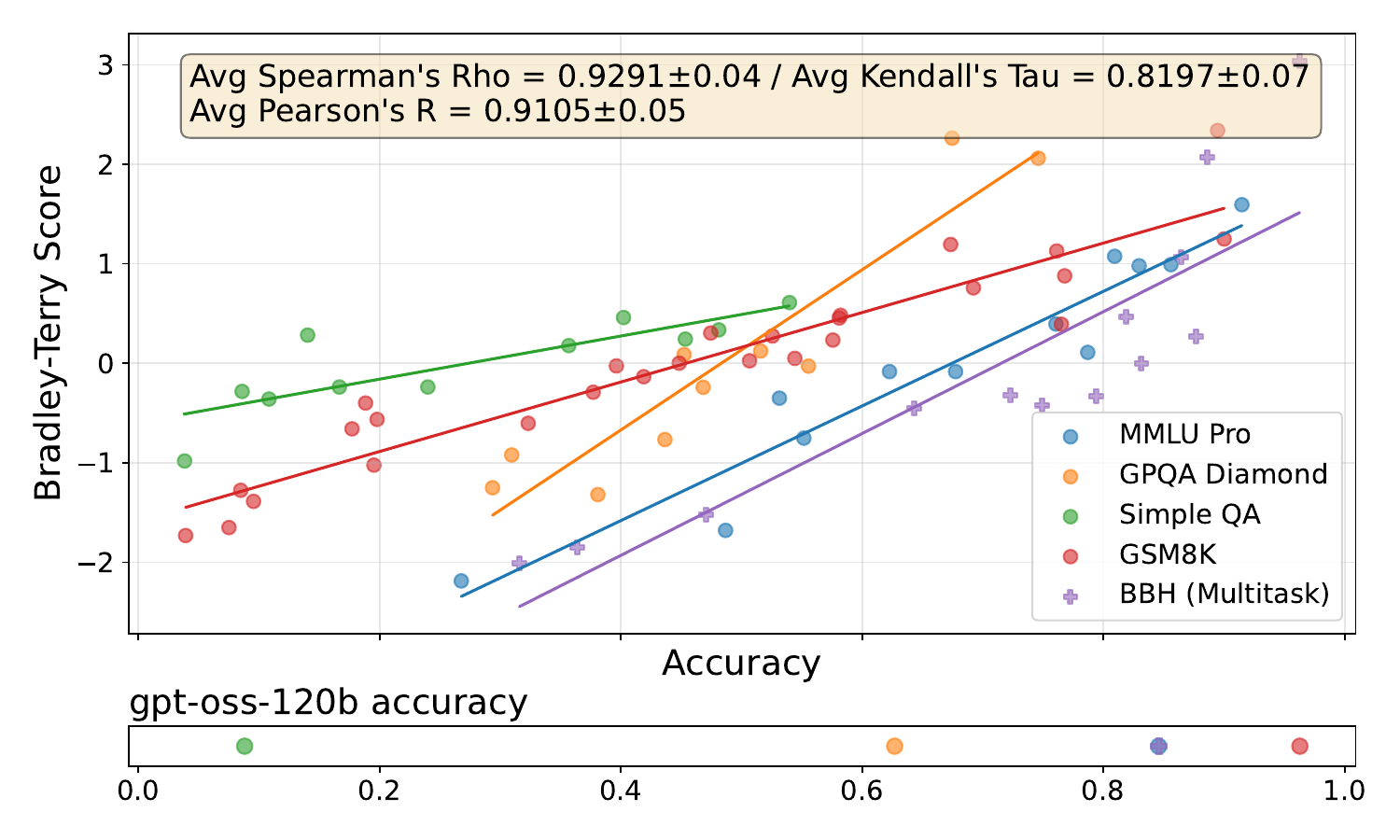}
        \caption{\texttt{gpt-oss-120b}}
        \label{fig:gpt-oss-120b}
    \end{subfigure}
    \begin{subfigure}{0.49\linewidth}
        \centering
    \includegraphics[width=\linewidth]{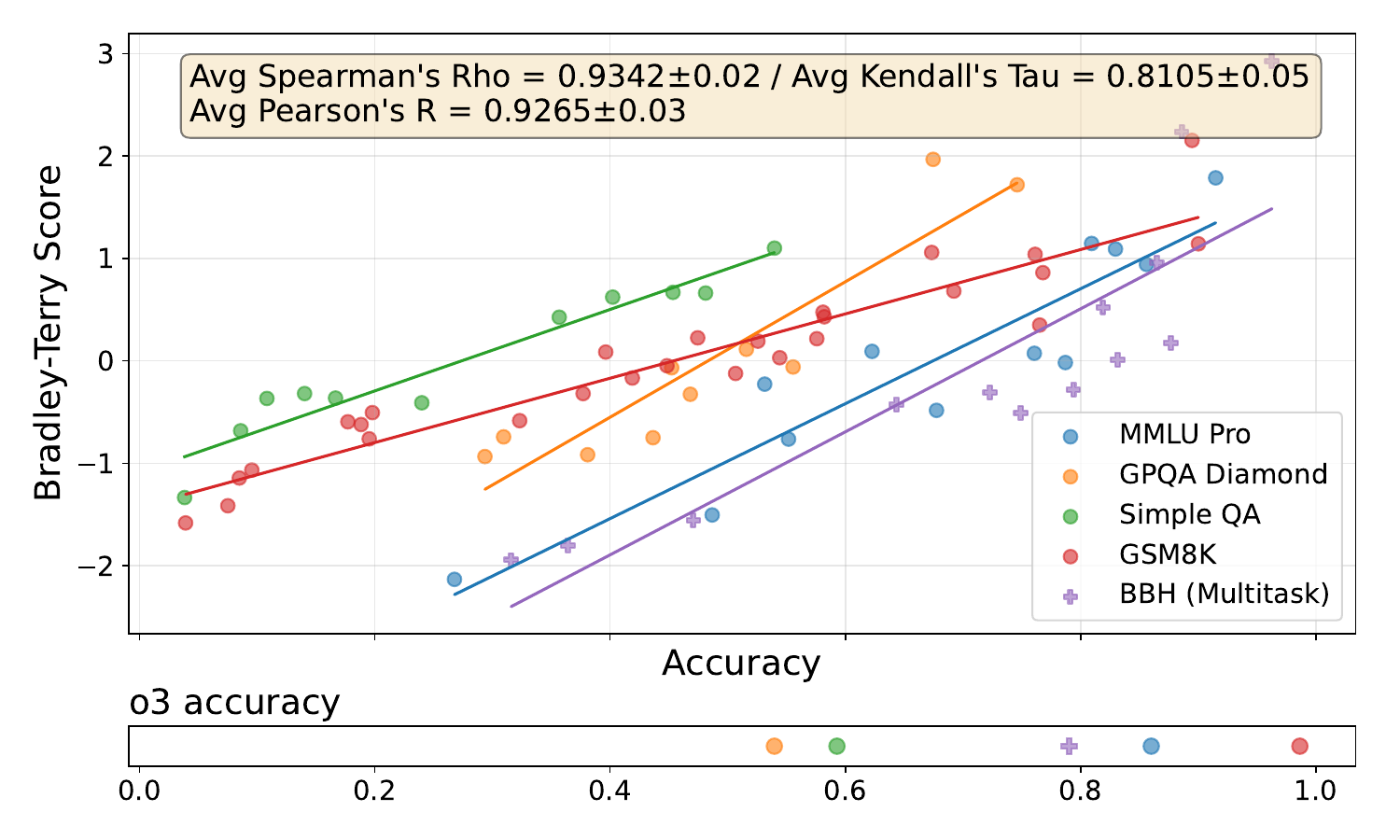}
    \caption{\texttt{o3}}
    \label{fig:o3}
    \end{subfigure}
    \caption{Correlation between accuracy and Bradley-Terry score using other models for collecting pairwise comparisons. Overall, we observe similarly impressive rank and score correlation for both \texttt{gpt-oss-120b} and \texttt{o3}.}
    \label{fig:other_models}
\end{figure}

% \begin{figure}[H]
%     \begin{subfigure}{0.49\linewidth}
%         \centering
%         \includegraphics[width=0.9\linewidth]{fig/rank_vs_score_correlation_gpt-oss-120b.pdf}
%         \caption{\texttt{gpt-oss-120b}}
%     \end{subfigure}
%     \begin{subfigure}{0.49\linewidth}
%         \centering
%         \includegraphics[width=0.9\linewidth]{fig/rank_vs_score_correlation_o3.pdf}
%         \caption{\texttt{o3}}
%     \end{subfigure}
%     \caption{Bradley-Terry is comparable to baseline.}
% \end{figure}

\subsection{Comparison to baseline on SimpleQA} \label{app:comp_to_baseline}

\begin{figure}[H]
    \centering
    \begin{subfigure}{0.49\linewidth}
        \centering
        \includegraphics[width=\linewidth]{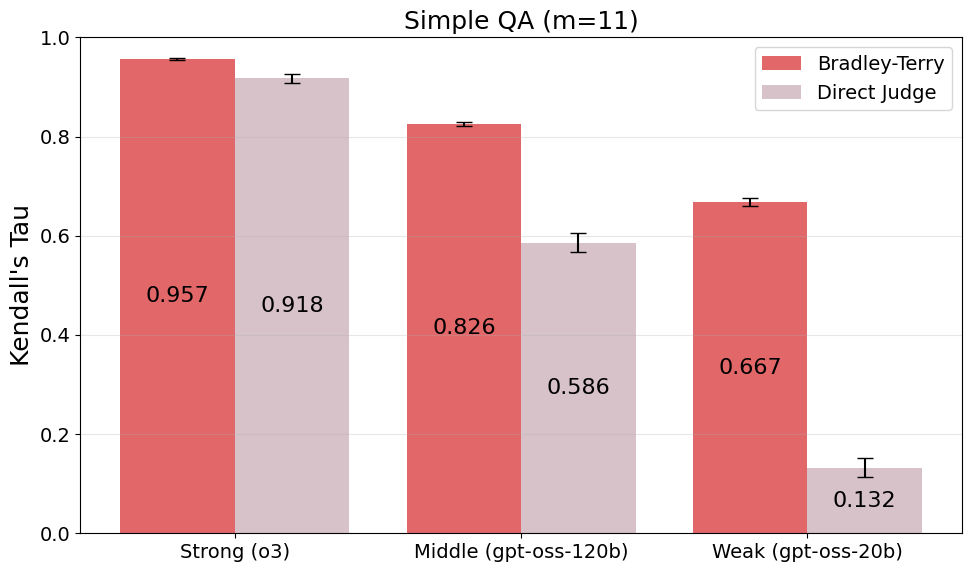}
        \caption{Rank correlation}
    \end{subfigure}
    \hfill
    \begin{subfigure}{0.49\linewidth}
        \centering
        \includegraphics[width=\linewidth]{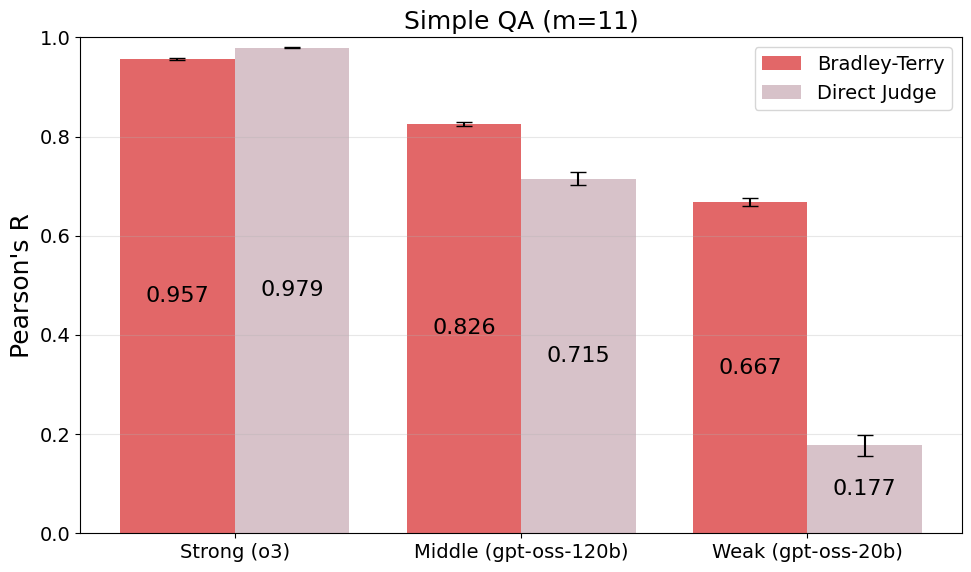}
        \caption{Score correlation}
    \end{subfigure}
    \caption{\textbf{Bradley-Terry vs. direct judge on Simple QA} We vary our evaluator between the ``\textit{strongest model}''  on the benchmark (\texttt{o3}, 59.3\% acc), a ``middle model'' (\texttt{gpt-oss-120b}, 8.8\% acc) and the ``\textit{weakest model}'' on the benchmark (\texttt{gpt-oss-20b}, 4.9\% acc). While the two methods are comparable using strong models, Bradley-Terry clearly dominates the baseline in the weaker model setting.}
    \label{fig:bt_vs_baseline_simpleqa_smw}
\end{figure}

\subsection{Controlling for bias} \label{app:bias}

\begin{figure}[H]
    \begin{subfigure}{0.49\linewidth}
        \centering
        \includegraphics[width=\linewidth]{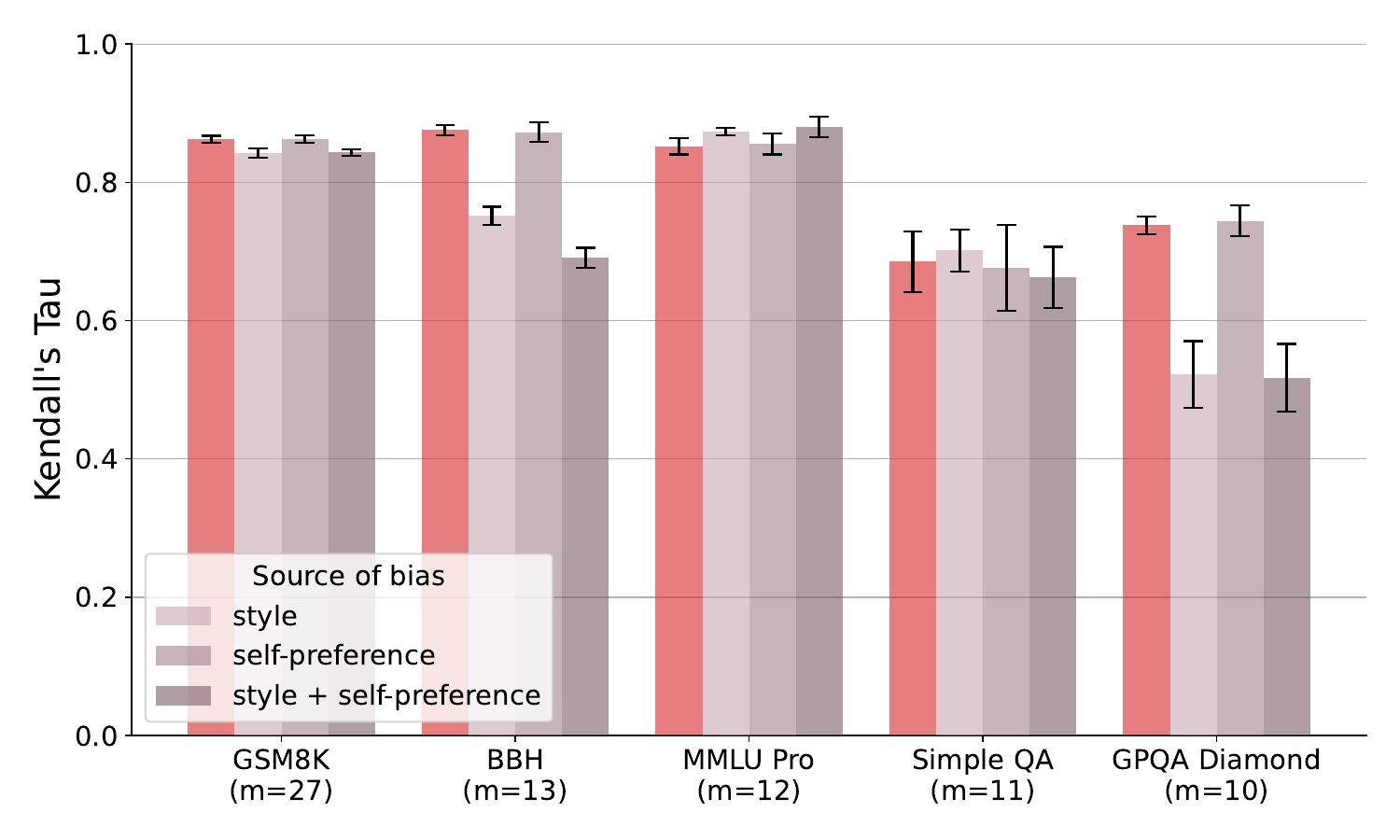}
        \caption{\texttt{gpt-oss-120b}}
    \end{subfigure}
    \begin{subfigure}{0.49\linewidth}
        \centering
        \includegraphics[width=\linewidth]{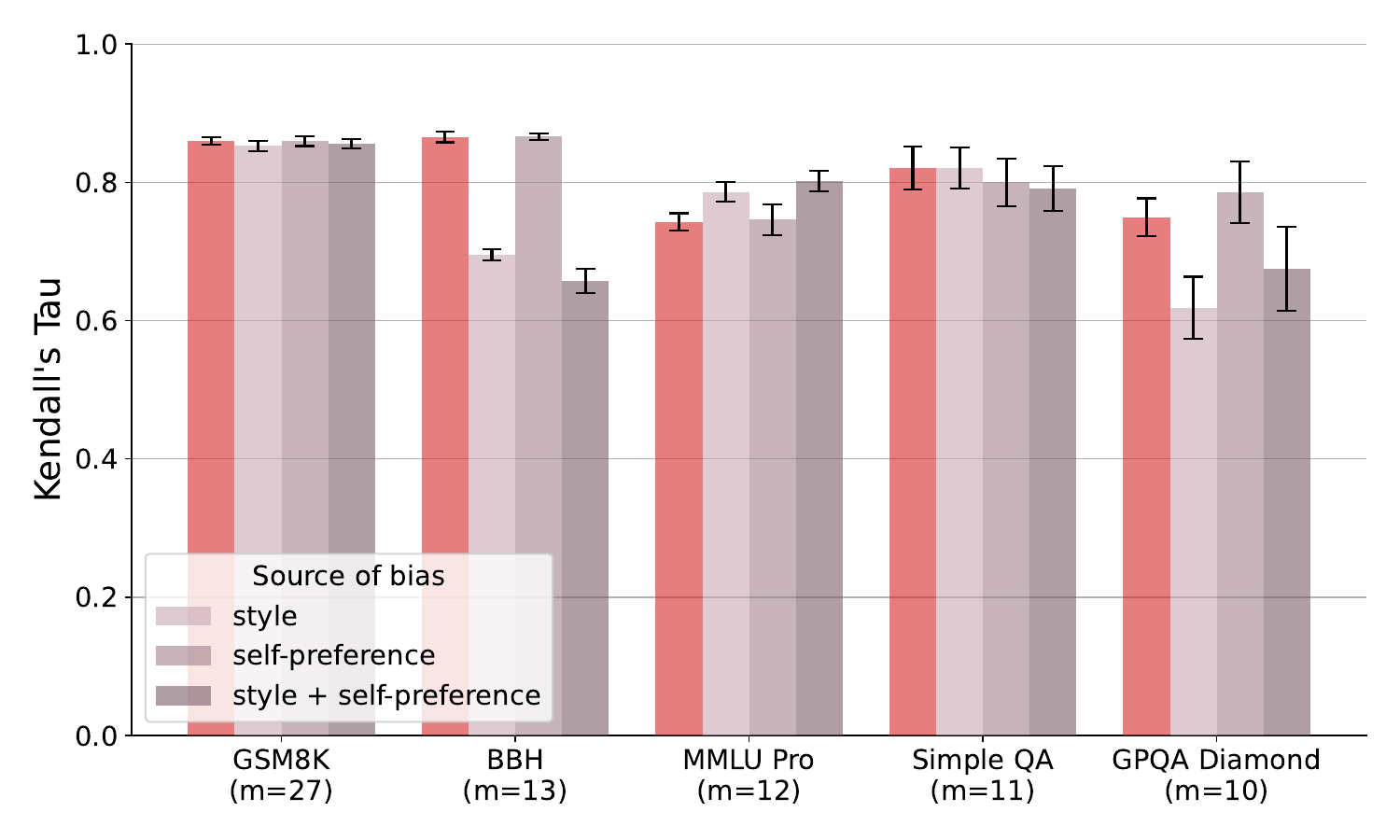}
        \caption{\texttt{o3}}
    \end{subfigure}
    \begin{subfigure}{0.49\linewidth}
        \centering
        \includegraphics[width=\linewidth]{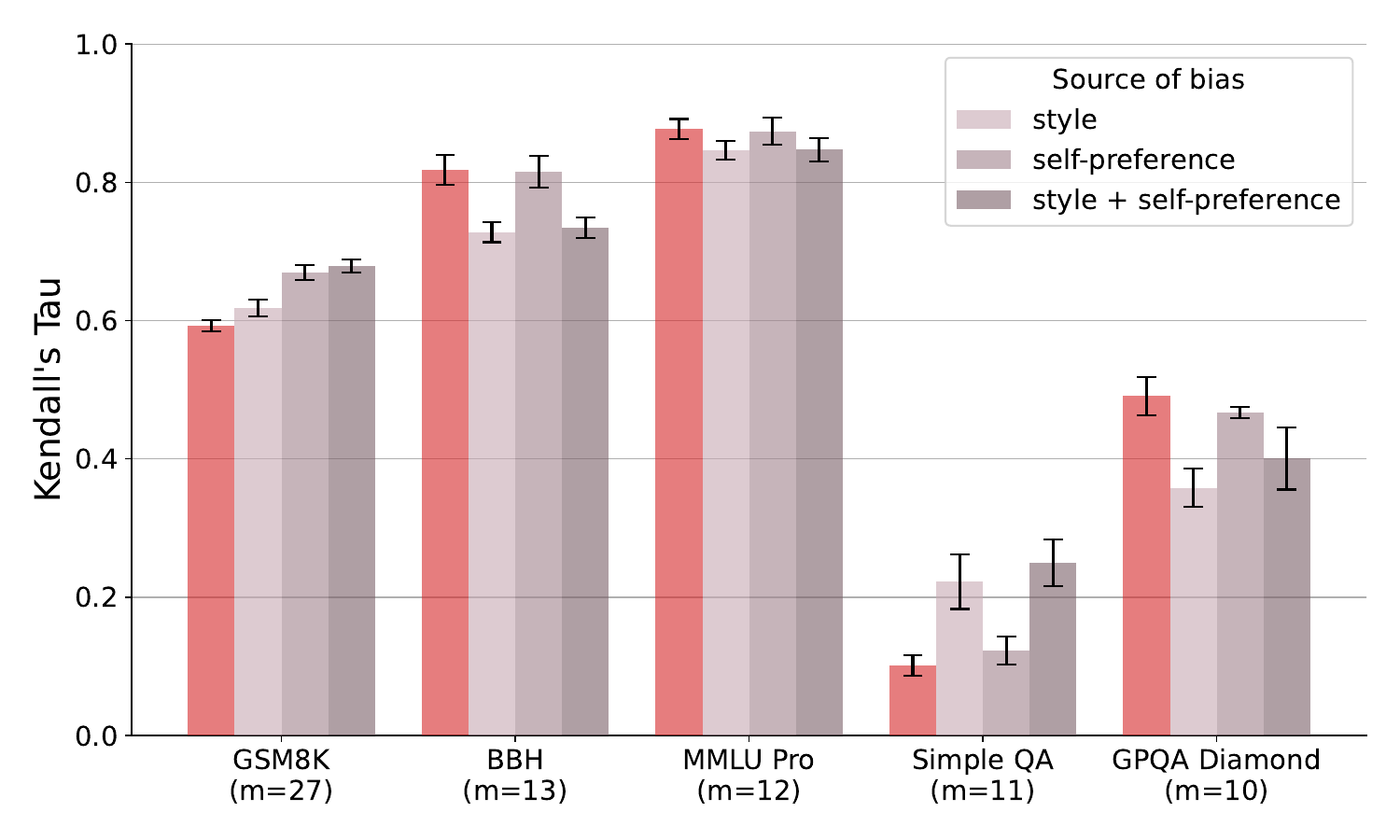}
        \caption{\texttt{phi-4}}
    \end{subfigure}
    \begin{subfigure}{0.49\linewidth}
        \centering
        \includegraphics[width=\linewidth]{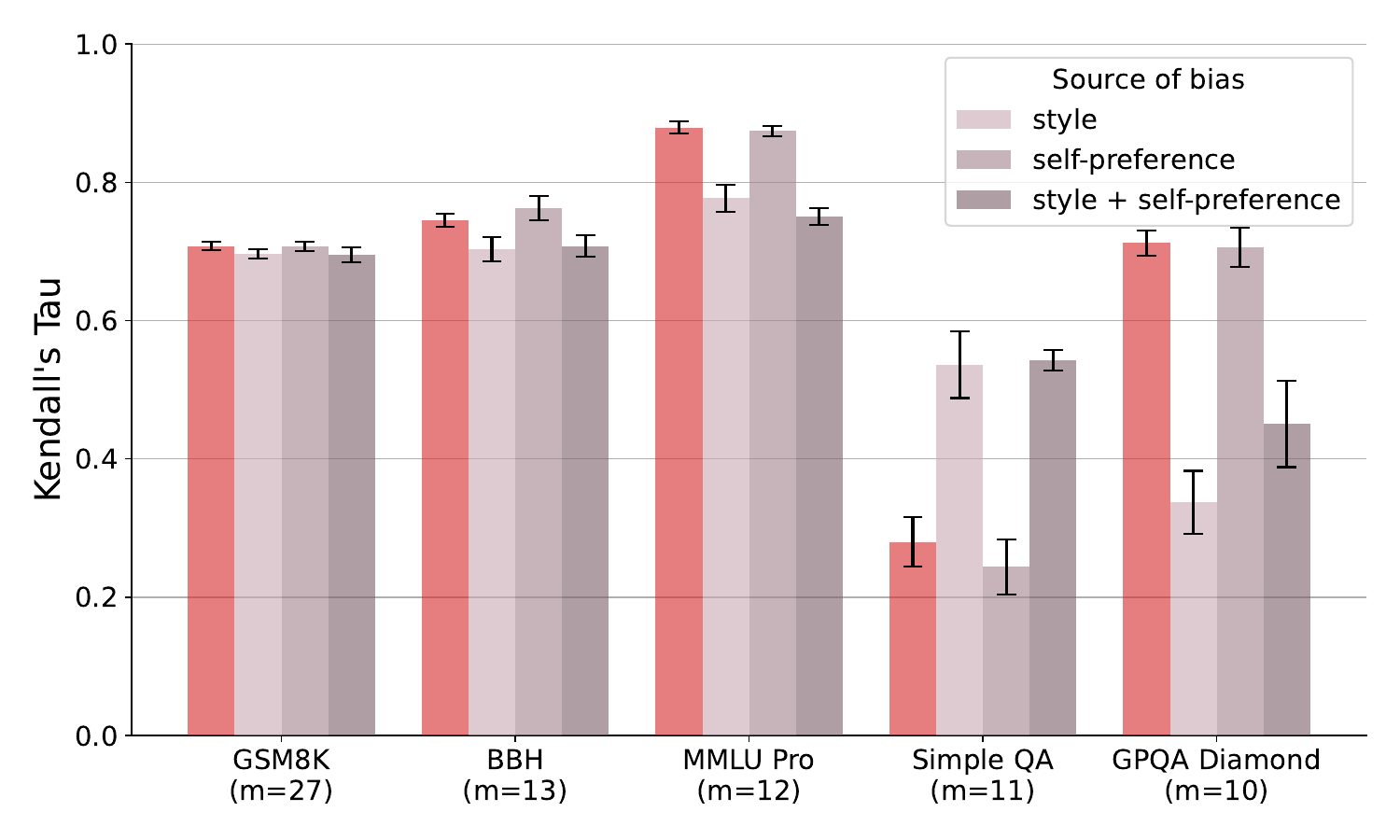}
        \caption{\texttt{gemma-3-27b-it}}
    \end{subfigure}
    \caption{Correcting for the most common type of judge biases. Original rank correlation (without bias correction) is \textcolor{customred}{red}. $m$ stands for number of ranked models. Error bars are 95\% confidence intervals on 100 bootstrap samples. \textit{Style} controls for judges' preferences for answer length and formatting. \textit{Self-preference} controls for preference towards answers generated by the same model family as the model itself.}
\end{figure}

\section{Non-discriminative and discriminative pairs} \label{app:filtered_pairs}

\begin{figure}[H]
    \begin{subfigure}{0.5\linewidth}
        \centering
        \includegraphics[width=\linewidth]{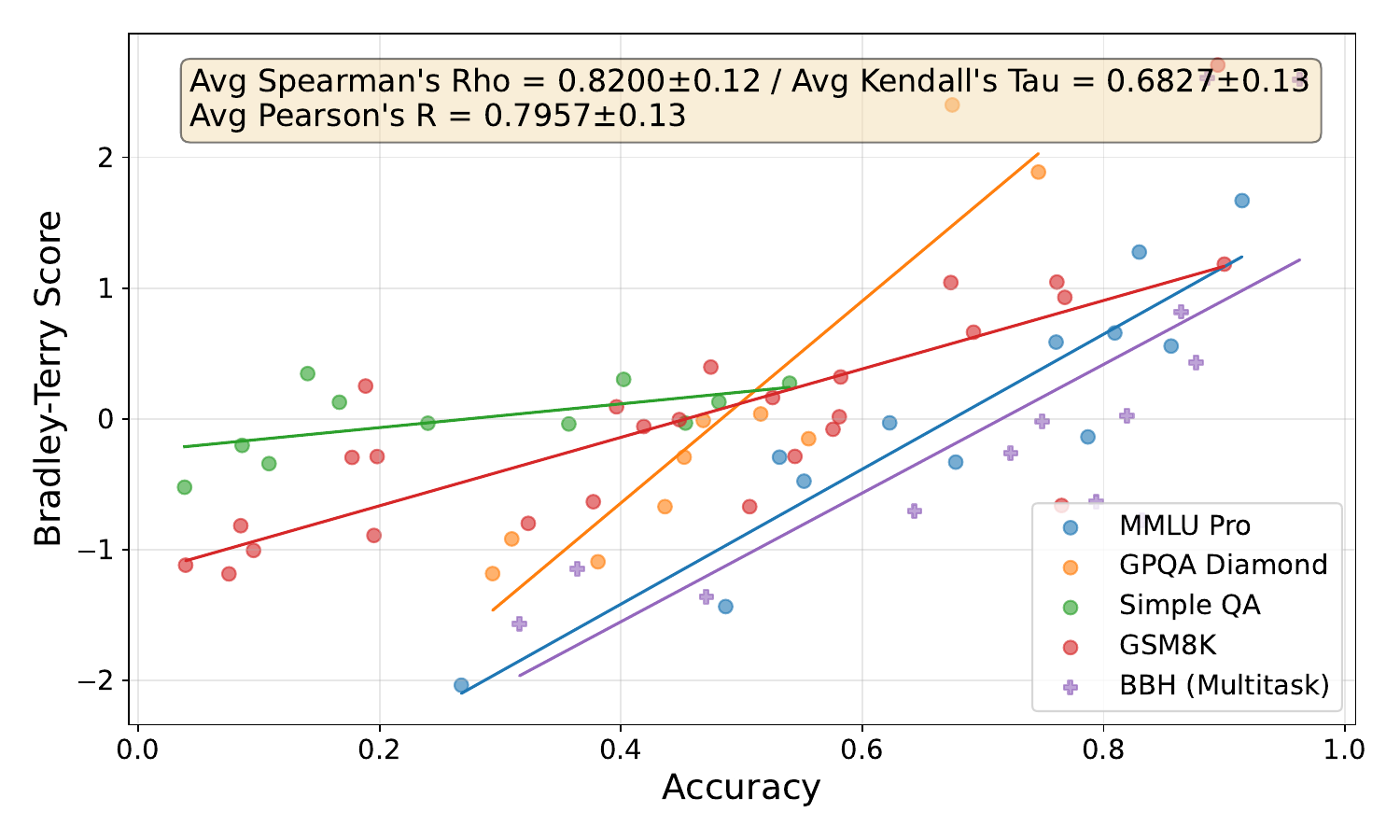}
        \caption{Non-discriminative pairs}
        \label{fig:cc_ii}
    \end{subfigure}
    \begin{subfigure}{0.5\linewidth}
        \centering
        \includegraphics[width=\linewidth]{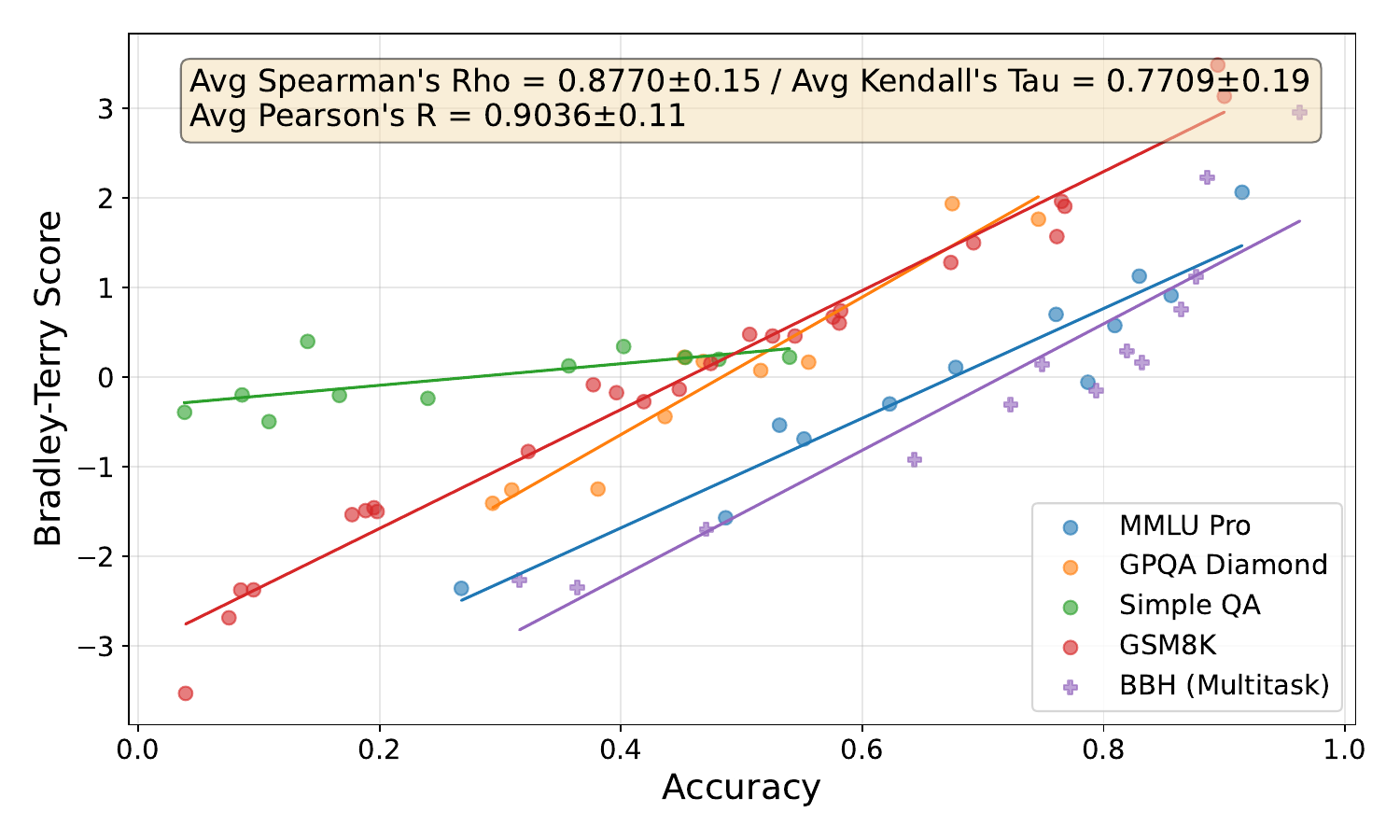}
        \caption{Discriminative pairs}
        \label{fig:ic_ci}
    \end{subfigure}
    \caption{\textbf{Results after filtering pairwise comparisons.} We test two settings: in one, we keep \textit{non-discriminative} pairs (both answers correct or both incorrect). This models the case where one of the distinguishing features is the appearance of the answer. Nevertheless, we still get surprisingly high rank correlation (Fig \ref{fig:cc_ii}). In the other, we keep \textit{discriminative} pairs (one answer correct, one incorrect), which improves rank correlation for most of our datasets over keeping all pairs (Fig \ref{fig:ic_ci}). Judge model: \texttt{gpt-oss-20b}}
    \label{fig:filtering}
\end{figure}

\section{Echo as a causal driver on non-discriminative pairs} \label{app:echo}
Echo is a failure mode in which a model, after producing its final answer, continues generating by repeating sequences. Echo can manifest as phrase repetition, or generating similar question-answer sequences following the original template.
We detected echo in pairwise comparisons on the BIG-Bench Hard (BBH) benchmark and found that it is strongly correlated with judge preference (Spearman $\rho \approx -0.50$) and with answer correctness.
This raised the question of whether echo causally drives judge decisions, or whether the correlation is mediated by correctness.

\subsection{Echo detection}
We detected the presence of echo in model responses using an LLM. We prompted the language model with a detailed description of what echo is, and provided examples and useful notes. We provide the full system prompt in Appendix \ref{app:echo-system-prompt}. We used \texttt{gpt-oss-120b} with 'Medium' reasoning effort and temperature 1.0. Across 13 models evaluated on BBH, 65\% of responses were flagged, though per-model rates varied substantially (from 0\% for GPT-4o to 97.8\% for Qwen3-0.6B).

\subsection{Causal analysis}
We ran a controlled intervention on BBH pairs where neither answer echoed, using these as the baseline.
For each pair, we built two counterfactuals by adding echo to one answer (appending the question--answer sequence three times), then queried \texttt{o3} on all three conditions.
Echo has a significant causal effect on judge preference (Figure~\ref{fig:echo_effect}): across 500 pairs per condition, adding echo to answer A drops $P(A>B)$ from 0.65 (original) to 0.29, while adding it to B raises $P(A>B)$ to 0.81 --- a contrast of $-0.52$ (95\% CI $[-0.57, -0.46]$).
This effect disappears on discriminative pairs --- those where exactly one answer is correct ($n=108$ per condition): the contrast shrinks to $-0.07$ (95\% CI $[-0.21, +0.06]$), which is indistinguishable from zero (Figure~\ref{fig:echo_effect_stratified}).
When ground truth is available, the judge relies on correctness rather than echo.

\section{Prompts}

\subsection{Direct judge prompt} \label{app:judge-no-GT}

\begin{lstlisting}[
  frame=single,
  basicstyle=\small\ttfamily,
  breaklines=true,
  breakindent=0pt,
  % caption={Prompt template for LLM-as-a-judge without ground truth},
  label=lst:judge-prompt,
  float=false
]
Your job is to look at a question and a predicted answer, and then assign a grade of either ["CORRECT", "INCORRECT", "NOT_ATTEMPTED"].
- CORRECT means that you think the answer is more likely correct.
- INCORRECT means that you think the answer is more likely incorrect.
- NOT ATTEMPTED means that there either is no response (empty string) or the answer is some form of "I don't know".
Simply reply with either CORRECT, INCORRECT, NOT ATTEMPTED. If the answer is attempted, you need to choose between CORRECT or INCORRECT. Even if you don't know the correct answer yourself, you have to guess to the best of your knowledge. Don't apologize or correct yourself if there was a mistake; we are just trying to grade the answer.
```
Question: {{question}}
Predicted answer: {{answer}}
```

Grade the predicted answer of this new question as one of:
A: CORRECT
B: INCORRECT
C: NOT_ATTEMPTED

Just return the letters "A", "B", or "C", with no text around it.
\end{lstlisting}

\subsection{Echo detection system prompt} \label{app:echo-system-prompt}

The echo detection system prompt can be broken down into the following sections:

\subsubsection{Summary}

\begin{lstlisting}[
  frame=single,
  basicstyle=\small\ttfamily,
  breaklines=true,
  breakindent=0pt,
  float=false
]
You will be presented a question and a model response. Your task is to determine whether the model response falls into one of two categories: "ECHO" or "NO_ECHO". Here are some examples of each category:
\end{lstlisting}

\subsubsection{Echo description}
\begin{lstlisting}[
  frame=single,
  basicstyle=\small\ttfamily,
  breaklines=true,
  breakindent=0pt,
  float=false
]
Model responses labeled "ECHO" represent a failure mode where the model answers the question, but then fails to stop generation. Examples of such failure modes include: 
- "Phrase / token echo": repeats the same phrase / token multiple times - typically until the model runs out of context. ("The answer is 42. The answer is 42. The answer" or "(4+5))))))))))")
- "Pattern echo": it's a form of sequence continuation, where the model generates NEW question answer pairs that are similar to the original question and answer. Often the original question-answer template is repeated. Even one pattern echo after the original question answer pair is considered an echo. ("Q: What is 2+2? A: 4. Q: What is 3+3? A: 6. Q: What is 4+4? A: 8.")
\end{lstlisting}

\subsubsection{No echo description}

\begin{lstlisting}[
  frame=single,
  basicstyle=\small\ttfamily,
  breaklines=true,
  breakindent=0pt,
  float=false
]
Model responses are NOT considered "ECHO" if:
- The model generates an answer to the question, and then stops. This includes cases where the model generates a correct answer, an incorrect answer, or a non-answer (e.g., "I don't know"). 
- The model arrives at a final answer by trial and error. The model may generate multiple answers, but reasons through them and picks one as its final answer.
- The final model answer is phrased in multiple ways. The model arrives at the final answer ("12+30=42") but then rephrases it ("The final answer is 42" or "#### 42"). This kind of rephrased repetition is not considered an echo.
- The reasoning process contains sub-questions related to the original question. The model may ask itself sub-questions as part of its reasoning process, but as long as it does not generate a new question-answer pair, it is not considered an echo.
- The model answer is verbose. The model may generate a long answer that includes a lot of information, but as long as it stops after generating its final answer, it is not considered an echo.
- The model starts reasoning, but runs out of tokens before it can generate a final answer. Such answers typically end with a truncated sentence or an incomplete reasoning process.
\end{lstlisting}

\subsubsection{Notes that help classification}
\begin{lstlisting}[
  frame=single,
  basicstyle=\small\ttfamily,
  breaklines=true,
  breakindent=0pt,
  float=false
]
Notes that will help you classify a model response:
1. Identify the model's final answer to the original question. Inspect the content after the final answer to determine if it contains failure modes typical of an echo.
2. Observe the original question-answer template - if it appears multiple times throughout the response, it is likely "ECHO".
3. If there is a new question-answer pair after the original question-answer pair, it is likely "ECHO".
4. If the model answer ends on a long sequence of repeated phrases or tokens, it is likely "ECHO".
\end{lstlisting}

\subsubsection{Examples}
\begin{lstlisting}[
  frame=single,
  basicstyle=\small\ttfamily,
  breaklines=true,
  breakindent=0pt,
  float=false
]
Example 1:

```
Q: What is the capital of France?
A: Paris.
```

This is "NO_ECHO" because the model generates a final answer ("Paris") and then stops.

Example 2:

```
Q: What is 2+2?
A: 4. Q: What is 3+3? A: 6. Q: What is 4+4? A: 8.
``` 

This is "ECHO" because the model generates new question-answer pairs after answering the original question.
\end{lstlisting}

\end{document}